\documentclass[11pt]{article}

\usepackage{ifthen}
\usepackage[normalem]{ulem} 
\usepackage{xcolor}
\usepackage{amssymb}

\newboolean{showedits}
\setboolean{showedits}{true} 
\ifthenelse{\boolean{showedits}}
{
\newcommand{\del}[1]{\textcolor{red}{\sout{#1}}} 
}{
\newcommand{\del}[1]{} 

}

\newboolean{showcomments}
\setboolean{showcomments}{true} 
\newcommand{\id}[1]{$-$Id: scgPaper.tex 32478 2010-04-29 09:11:32Z oscar $-$}

\ifthenelse{\boolean{showcomments}}
{\newcommand{\nbc}[3]{
{\colorbox{#3}{\bfseries\sffamily\scriptsize\textcolor{white}{#1}}}
{\textcolor{#3}{\sf\small$\blacktriangleright$\textit{#2}$\blacktriangleleft$}}}
}
{\newcommand{\nbc}[3]{}
\renewcommand{\del}[1]{} 
}

\definecolor{ibcolor}{rgb}{0.9,0.5,0}
\definecolor{dsrcolor}{rgb}{0.5,0.6,0}
\definecolor{cfcolor}{rgb}{0,0.5,0.9}
\definecolor{lwcolor}{rgb}{0.2,0.8,0.4}
\definecolor{eycolor}{rgb}{0.7,0.6,1.0}
\definecolor{oldcolor}{rgb}{0.2,0.2,0.2}
\definecolor{tdcolor}{rgb}{0.0,0.5,0.7}

\newcommand{\beq}{\vspace{0mm}\begin{equation}}
\newcommand{\eeq}{\vspace{0mm}\end{equation}}
\newcommand{\beqs}{\vspace{0mm}\begin{eqnarray}}
\newcommand{\eeqs}{\vspace{0mm}\end{eqnarray}}
\newcommand{\barr}{\begin{array}}
\newcommand{\earr}{\end{array}}

\newcommand{\pv}[0]{{\boldsymbol{p}}}

\newcommand{\zv}{\boldsymbol{z}}

\newcommand{\phiv}{\boldsymbol{\phi}}

\usepackage{booktabs}
\usepackage{array}
\newcolumntype{L}[1]{>{\raggedright\let\newline\\\arraybackslash\hspace{0pt}}m{#1}}
\newcolumntype{C}[1]{>{\centering\let\newline\\\arraybackslash\hspace{0pt}}m{#1}}
\newcolumntype{R}[1]{>{\raggedleft\let\newline\\\arraybackslash\hspace{0pt}}m{#1}}

\usepackage{acl}
\usepackage{bm}

\usepackage{times}
\usepackage{latexsym}

\usepackage[T1]{fontenc}
\usepackage[utf8]{inputenc}
\usepackage{microtype}
\usepackage{inconsolata}

\usepackage{graphicx}
\usepackage{booktabs}
\usepackage{placeins}
\usepackage{multirow}
\usepackage{amsmath,amssymb}
\usepackage{algorithm}
\usepackage{algpseudocode}
\usepackage{xcolor}
\usepackage[table]{xcolor} %

\providecommand{\bestval}[1]{\mathbf{#1}}
\providecommand{\secondval}[1]{\underline{#1}}
\providecommand{\best}[1]{\ensuremath{\mathbf{#1}}}
\providecommand{\second}[1]{\underline{#1}}
\providecommand{\uparrowcenter}{\raisebox{0.15ex}{$\uparrow$}}
\providecommand{\downarrowcenter}{\raisebox{0.15ex}{$\downarrow$}}

\title{Bayesian Sparse Low-Rank Adaptation for Large Language Model Uncertainty Estimation}

\author{
\textbf{Jijie Zhang}\textsuperscript{1},
\textbf{Zhe Ren}\textsuperscript{1},
\textbf{Quan Zhang}\textsuperscript{2}\thanks{Corresponding authors},
\textbf{Dandan Guo}\textsuperscript{1,3}\footnotemark[1]
\\[1ex]
\textsuperscript{1}School of Artificial Intelligence, Jilin University
\\
\textsuperscript{2}Broad College of Business, Michigan State University
\\
\textsuperscript{3}Center of Excellence for Generative AI, King Abdullah University of Science and Technology (KAUST)
\\
\texttt{quan.zhang@broad.msu.edu},
\texttt{guodandan@jlu.edu.cn}
}
  
\begin{document}
\maketitle

\renewcommand{\thefootnote}{}
\footnotetext{To appear in EMNLP 2026.}
\renewcommand{\thefootnote}{\arabic{footnote}}

\begin{abstract}

Large language models (LLMs) exhibit remarkable reasoning capabilities, but their task-specific fine-tuning is notoriously plagued by overconfidence, severely hindering trustworthy deployment. We propose Data-Adaptive Lower-Rank Adaptation (DALorRA), a simple and effective variational Bayesian sparse framework that shifts the paradigm of uncertainty quantification from the dense parameter space to the lightweight rank level of low-rank adaptation (LoRA). With the insight that LoRA essentially aggregates multiple rank-one components that may provide superfluous model capacity,  DALorRA imposes stochastic masking on rank dimensions, enabling Bayesian regularization of model capacity during training and ensemble-like calibration during inference. Extensive experiments demonstrate DALorRA's excellent calibration of LLMs without compromising reasoning accuracy. Code is available at \url{https://github.com/John-Wilson-ML/DALorRA}.
\end{abstract}

{\section{Introduction}}

Large language models (LLMs) exhibit remarkable reasoning capabilities, but their reliability remains a central obstacle to trustworthy deployment \citep{xiong2024can,shelmanov2025uncertainty,bouchard2026uqlm}. When adapted to downstream tasks via deterministic fine-tuning, LLMs can memorize the target distribution, yielding overconfident predictions even for incorrect answers \citep{yang2024bayesian,li2025calibrating}. This miscalibration weakens the interpretability of predictive confidence and highlights an urgent need for scalable uncertainty quantification. While Bayesian neural networks and Deep Ensembles provide rigorous uncertainty quantification \citep{blundell2015weight,lakshminarayanan2017simple}, performing Bayesian inference over billion-parameter weight spaces or maintaining multiple model copies is computationally prohibitive for LLMs \citep{nemani2025efficient,xiang2026scalable}.

Parameter-efficient fine-tuning, particularly Low-Rank Adaptation \citep[LoRA,][]{hu2022lora}, circumvents full weight updates and has emerged as a promising foundation for scalable uncertainty quantification \citep{wang2024blob,shi2026training}. However, LoRA-based calibration largely focuses on the uncertainty of the LoRA adapter, incurring significant computational overhead and compromising the lightweight nature and efficiency that make LoRA so appealing \citep{marszalek2025minimal}.  Moreover, these approaches predominantly face a structural bottleneck: The LoRA rank $r$ is treated as a static, globally fixed hyperparameter \citep{gu2025lora,zhou2026lara}. Since the true intrinsic dimensionality varies across downstream tasks, rigidly fixing~$r$ can introduce superfluous model capacity that exacerbates overfitting in low-data regimes \citep{zhang2023adalora,liu2024alora,cheon2026brain}. As frontier LLMs tend to require minimal parameter adjustments for specialized tasks, there is a growing need for adaptation that can dynamically prune unnecessary rank components without sacrificing core capabilities ~\citep{huang2025dynamic,kumaravelu2026post,deng2026dr}.

In this work, we circumvent the computational difficulty of Bayesianizing the full LoRA adapter and relax the fixed LoRA rank by directly modeling the rank uncertainty. Given rank $r$, LoRA essentially utilizes the combination of $r$ rank-one components that may provide superfluous capacity. We propose \underline{D}ata-\underline{A}daptive \underline{Lo}we\underline{r}-\underline{R}ank \underline{A}daptation (DALorRA). Specifically, we inject a stochastic diagonal mask matrix $\mathbf{D} \in \mathbb{R}^{r \times r}$, formulating the weight update as $\Delta \mathbf{W} = \mathbf{B}\mathbf{D}\mathbf{A}$. By treating the diagonal elements as latent variables governed by learnable Bernoulli distributions via variational inference, DALorRA captures discrete structural uncertainty over which adaptation dimensions should be active. As $\mathbf{D}$ deactivates components of $\mathbf{B}$ and $\mathbf{A}$ with deactivation probabilities depending on downstream tasks, we are essentially implementing a lower-rank adaptation data-adaptively. 

Despite its simplicity, DALorRA achieves outstanding performance in uncertainty quantification and reasoning accuracy by bridging Bayesian neural networks and deep ensembles in a LoRA framework. Analogous to Bayesian neural networks, DALorRA learns the posterior distribution over the mask $\mathbf{D}$,  capturing epistemic uncertainty and pruning unnecessary model complexity to avoid overfit. It also mirrors a deep ensemble by stochastically aggregating the predictions of diverse rank-one configurations, maintaining outstanding prediction accuracy and high efficiency. More intuitive justifications are deferred to the end of Section~\ref{sec:method}.
We summarize three primary contributions:
\begin{itemize}
    \item  

\textbf{Methodology innovation.} We shift the paradigm of uncertainty quantification from the dense LLM parameter space or LoRA adapter ($\mathbf{B}$ and/or $\mathbf{A}$) space to the lightweight LoRA rank level, introducing minimal computational overhead. 
   \item 
    \textbf{Principled Approach.} We bridge theoretically rigorous variational Bayesian estimation with empirically sound ensemble-like inference. This principled approach empirically outperforms rank-level dropout, which similarly induces uncertainty and sparsity over the LoRA rank. 
    
    \item 
 {\textbf{Comprehensive empirical validation and excellent performance.}} We demonstrate through extensive experiments on diverse reasoning benchmarks that DALorRA consistently achieves exceptional calibration without sacrificing reasoning accuracy.
\end{itemize}
\begin{figure*}[t]
    \centering
    \includegraphics[width=0.95\textwidth]{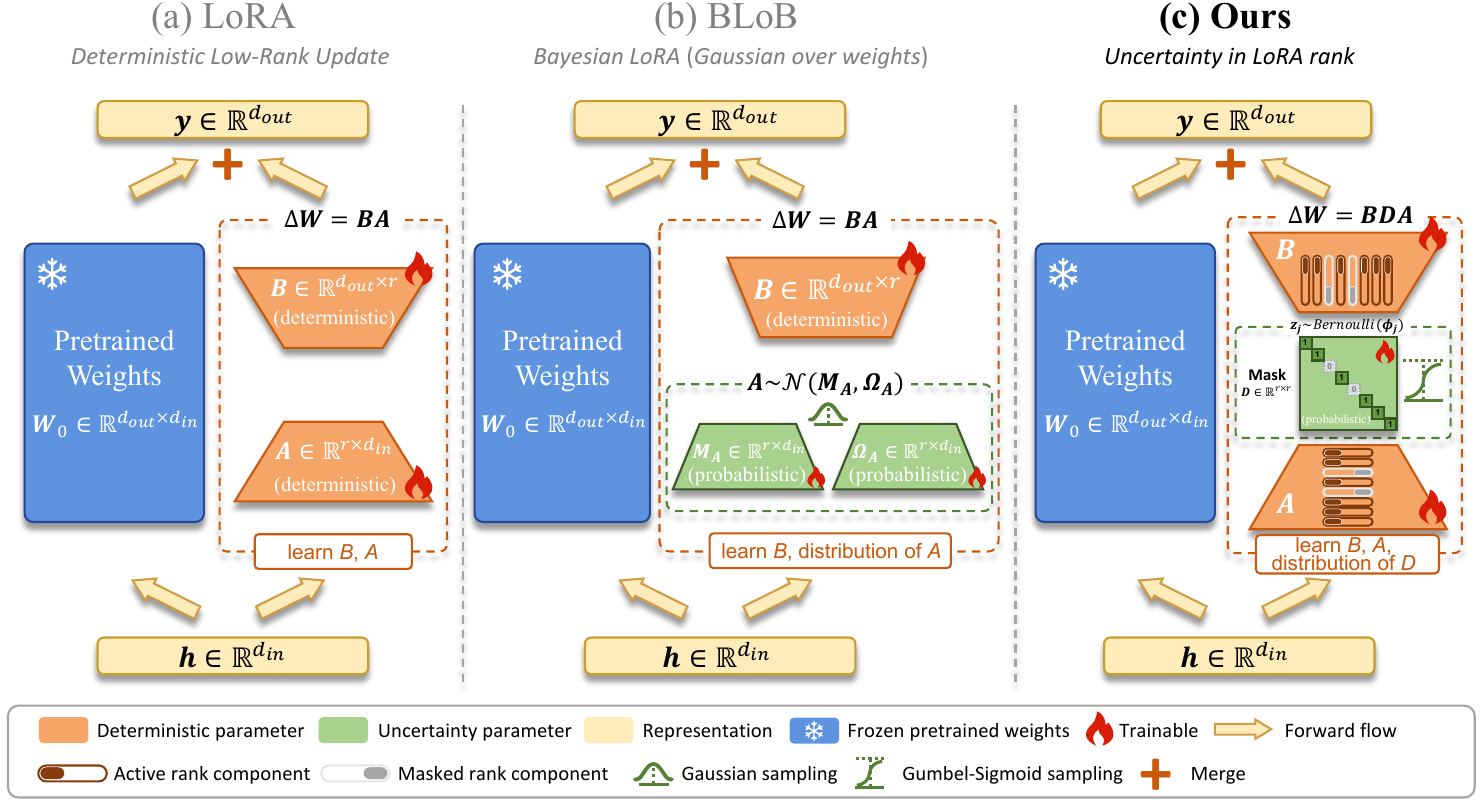}
    \caption{
    {Comparison between LoRA, BLoB, and our DALorRA.}
    Standard LoRA learns a deterministic low-rank update $\Delta \mathbf{W}=\mathbf{BA}$.
    BLoB models Gaussian uncertainty over LoRA adapter $\mathbf{A}$.
    In contrast, DALorRA introduces a stochastic diagonal mask
    $\mathbf{D}$, yielding $\Delta W = BDA$ to model rank-level uncertainty in LoRA.}
    \label{fig:masklora_overview}
    \vspace{-0.5em}
\end{figure*}

\section{Related Work}
We review related work in three streams: LLM output uncertainty, rank-adaptive LoRA and dropout-based inference for LLMs, and LLM uncertainty via parameter-efficient fine-tuning. The proposed DALorRA contributes to the second and third streams.

\subsection {LLM Output Uncertainty}
Reliable uncertainty estimation is critical for mitigating hallucinations and overconfident predictions in large language models \citep{shelmanov2025uncertainty,bouchard2026uqlm}. Existing research has approached this from the output space, utilizing heuristics such as verbalized confidence \citep{lin2022teaching}, prompt-based ensembling \citep{jiang2023calibrating}, and semantic entropy across multiple generations \citep{kuhn2023semantic,shelmanov2025uncertainty,bouchard2026uqlm}. While highly appealing due to their plug-and-play nature, output-space metrics inherently bypass the underlying probability distributions of model parameters. Consequently, when adapting LLMs to specialized downstream tasks in low-data regimes, such output-centric heuristics do not fully capture epistemic uncertainty during fine-tuning, leaving models prone to overconfident predictions and overfitting~\citep{niu2024functional,bakman2025uncertainty}. This limitation necessitates a shift toward theoretically grounded, parameter-level uncertainty quantification during fine-tuning. Complementary training-time calibration methods for conventional DNNs improve
confidence--accuracy alignment through classifier design
\citep{ni2025balancing} or adaptive stochastic masking that reduces classifier shift \citep{ni2026deep}.

\subsection{Rank-Adaptive LoRA and Dropout-Based Approximate Inference}
{
Rank-adaptive LoRA methods improve parameter efficiency by learning, pruning,
or reallocating low-rank adaptation capacity. SoRA learns continuous rank gates
and induces sparsity through $\ell_1$ regularization and proximal optimization
\citep{ding2023sparse}. AdaLoRA and ALoRA allocate rank budgets according to
importance estimates, whereas DyLoRA trains adapters that support multiple
selectable ranks \citep{zhang2023adalora,liu2024alora,valipour2023dylora}.
Related approaches allocate heterogeneous ranks across experts in
mixture-of-experts models or Transformer layers, including DR-LoRA, LaRA, and
La-LoRA \citep{deng2026dr,zhou2026lara,gu2025lora}. Another line of work
studies dropout. Monte Carlo (MC) dropout retains
Bernoulli dropout masks at test time and estimates predictive uncertainty
through MC sampling \citep{gal2016dropout}; variational dropout
interprets multiplicative Gaussian noise as variational inference over
weights, while sparse variational dropout learns dropout rates to induce
weight-level sparsity \citep{kingma2015variational,molchanov2017variational}.
At the LoRA rank level, DropLoRA applies a fixed dropout rate during training; during inference, it disables dropout and uses weight scaling \citep{zhang2025droplora}.

Our work advances this line of research through a principled Bayesian framework for LoRA rank selection and LLM uncertainty quantification. Although SoRA and DropLoRA also induce rank-level sparsity, they differ from DALorRA in model formulation, estimation, and intended use; both target adaptive rank selection to improve reasoning accuracy rather than calibration. SoRA learns continuous diagonal gates with an $\ell_1$ penalty. Its lasso-like gate regularization makes uncertainty quantification intrinsically difficult \citep{leeb2005model,tibshirani2016exact,hoppe2024non}, even when employing weight-level uncertainty.  DropLoRA fixes or tunes a dropout rate during training and disables dropout at inference, yielding a deterministic update equivalent to standard LoRA. In contrast, DALorRA inherently supports uncertainty quantification through Bayesian inference while remaining lightweight and achieving strong predictive performance.

\subsection{Uncertainty via Parameter-Efficient Fine-Tuning}
Classical uncertainty estimation methods, such as Monte Carlo Dropout \citep{gal2016dropout} and Deep Ensembles \citep{lakshminarayanan2017simple}, approximate predictive uncertainty through stochastic inference or model averaging, but can be computationally prohibitive when directly applied to LLMs. 
To avoid performing uncertainty estimation over the full parameter space,
parameter-efficient fine-tuning (PEFT), particularly LoRA \citep{hu2022lora}, has emerged as a practical foundation for uncertainty-aware adaptation.

Recent work has investigated how to Bayesianize or calibrate only the lightweight PEFT modules rather than the full LLM. Post-hoc Laplace approximation has been applied to fine-tuned LoRA parameters to obtain approximate posterior uncertainty \citep{wang2024blob,shi2026training,marszalek2025minimal,rahmati2026c}. Advancing this paradigm towards end-to-end inference, Bayesian 
low-rank adaptation by backpropagation \citep[BLoB,][]{wang2024blob} moves posterior learning directly into the fine-tuning stage by jointly optimizing the variational mean and covariance of LoRA adapters. Training-Free Bayesianization \citep[TFB,][]{shi2026training} further converts trained LoRA adapters into Bayesian ones by searching over low-rank isotropic Gaussian posteriors. More recently, Contextual LoRA  \citep[C-LoRA,][]{rahmati2026c} introduces modules to capture sample-dependent uncertainty to complement global parameter uncertainty. Complementary to uncertainty estimation, safe fine-tuning methods seek to preserve safety alignment during adaptation through safety-sensitive subspace constraints or distributional data alignment \citep{zhang2026guardrail,wang2026safeguarding}.

Our work contributes to this stream of uncertainty quantification but is distinct from extant research, which largely estimates the variances of LoRA adapters $\mathbf{B}$ and/or $\mathbf{A}$, potentially compromising LoRA's efficiency. Instead, the proposed DALorRA targets uncertainty at the rank level, introducing negligible computational overhead. Furthermore, our stochastic masking configuration imposes explicit sparsity on the already lightweight LoRA architecture, making it particularly suitable for frontier LLMs that require minimal adjustments and thus lower-rank adaptation for specialized tasks. Ultimately, our method bridges theoretically rigorous Bayesian neural networks with empirically robust deep ensembles, which underpin its excellent performance. We compare DALorRA with regular LoRA and BLoB in Figure~\ref{fig:masklora_overview}; instead of learning the uncertainty of LoRA adapters, we keep them deterministic and learn the posterior distribution of the rank mask~$\mathbf{D}$.

{\section{Preliminaries}}

\subsection{Low-Rank Adaptation}
Low-Rank Adaptation (LoRA) is a parameter-efficient fine-tuning paradigm that adapts a pre-trained model by freezing the original dense weights and optimizing a low-rank update \citep{hu2022lora}. Given a linear layer with a frozen pretrained weight $\mathbf{W}_0\in\mathbb{R}^{d_{\mathrm{out}}\times d_{\mathrm{in}}}$, LoRA parameterizes the task-specific update as $\Delta\mathbf{W}=\mathbf{B}\mathbf{A}$, where $\mathbf{B}\in\mathbb{R}^{d_{\mathrm{out}}\times r}$, $\mathbf{A}\in\mathbb{R}^{r\times d_{\mathrm{in}}}$, and $r\ll\min(d_{\mathrm{out}},d_{\mathrm{in}})$ is the adaptation rank. For an input representation $\mathbf{h}\in\mathbb{R}^{d_{\mathrm{in}}}$, the forward pass is
\begin{equation*}
    \mathbf{o} = (\mathbf{W}_0+\mathbf{B}\mathbf{A})\mathbf{h},
\end{equation*}
where $\mathbf{o}\in\mathbb{R}^{d_{\mathrm{out}}}$ is the output representation. Compared to full-parameter fine-tuning, LoRA introduces only $r\times(d_{\mathrm{out}}+d_{\mathrm{in}})$ trainable parameters, substantially reducing memory and computational costs while largely preserving adaptation efficacy.

\subsection{Confidence Calibration in LLMs}
Following standard protocols \citep{wang2024blob,yang2024bayesian}, we consider a downstream dataset $\mathcal{D}=\{(\mathbf{x}_i,y_i)\}_{i=1}^{N}$ and formulate LLM reasoning and multiple-choice tasks as next-token classification over a candidate label set $\mathcal{Y}$. Given an input prompt $\mathbf{x}$, the LLM (parameterized by $\boldsymbol{\theta}$) yields a predictive distribution $p_{\boldsymbol{\theta}}(y \mid \mathbf{x})$. The predicted label and its confidence are 

\begin{equation*}
\hat{y} = \arg\max_{y\in\mathcal{Y}} p_{\boldsymbol{\theta}}(y\mid \mathbf{x}), \quad \hat{p} = \max_{y\in\mathcal{Y}} p_{\boldsymbol{\theta}}(y\mid \mathbf{x}).
\end{equation*}

When adapted to downstream tasks via deterministic fine-tuning, LLMs frequently overfit the target distribution, yielding overconfident predictions even for incorrect answers. To quantify this miscalibration, Expected Calibration Error (ECE) partitions predictions into $M$ bins $\{B_m\}_{m=1}^{M}$:
\begin{equation*}
    \mathrm{ECE} = \sum_{m=1}^{M} \frac{|B_m|}{N} \left| \mathrm{acc}(B_m) - \mathrm{conf}(B_m) \right|,
\end{equation*}
where $\mathrm{acc}(B_m) = \frac{1}{|B_m|} \sum_{i \in B_m} \mathbb{I}(\hat{y}_i = y_i)$ and $\mathrm{conf}(B_m) = \frac{1}{|B_m|} \sum_{i \in B_m} \hat{p}_i$, with $\mathbb{I}(\cdot)$ being the indicator function. This severe overconfidence phenomenon necessitates a paradigm shift toward principled uncertainty quantification.

\subsection{Variational Inference}
Bayesian neural networks mitigate overconfidence by placing prior distributions over weights to capture epistemic uncertainty \citep{blundell2015weight}. Since computing the exact posterior $p(\mathbf{W}\mid\mathcal{D})$ is intractable, Variational Inference (VI) approximates it with a tractable distribution $q_{\boldsymbol{\phi}}(\mathbf{W})$ parameterized by variational parameters $\boldsymbol{\phi}$.

The optimal $\boldsymbol{\phi}$ is obtained by minimizing the Kullback-Leibler (KL) divergence to the true posterior, which is mathematically equivalent to minimizing the negative Evidence Lower Bound:
\begin{equation*}
    \label{eq:general_elbo}
    -\mathbb{E}_{q_{\boldsymbol{\phi}}} \left[ \log p(\mathcal{D}\mid\mathbf{W}) \right] + \mathrm{KL} \left( q_{\boldsymbol{\phi}}(\mathbf{W}) \,\|\, p(\mathbf{W}) \right).
\end{equation*}
The first term corresponds to the expected negative log-likelihood, and the second term regularizes the deviation of $q_{\boldsymbol{\phi}}(\mathbf{W})$ from the prior $p(\mathbf{W})$. To enable end-to-end backpropagation through the expectation term, reparameterization tricks \citep{kingma2013auto} are often required. For instance, if $q_{\boldsymbol{\phi}}$ is Gaussian with $\boldsymbol{\phi}=(\boldsymbol{\mu},\boldsymbol{\Sigma})$, a weight sample is drawn via $\mathbf{W} = \boldsymbol{\mu} + \boldsymbol{\Sigma}^{1/2}\boldsymbol{\epsilon}$, where $\boldsymbol{\epsilon} \sim \mathcal{N}(\mathbf{0},\mathbf{I})$ is independent of $\phiv$. This framework lays the theoretical foundation for our method.

\section{Method}
\label{sec:method}
 
We propose Data-Adaptive Lower-Rank Adaptation (DALorRA), a variational Bayesian sparse framework tailored for uncertainty-aware LLM fine-tuning. Unlike LoRA (Figure \ref{fig:masklora_overview}a) that relies on fixed dense updates, or recent Bayesian extensions, like BLoB (Figure \ref{fig:masklora_overview}b), which may suffer from high optimization overhead of Gaussian posteriors over the dense LoRA adapters, our approach represents a structural paradigm shift. Our core insight is that the standard LoRA update matrix $\Delta\mathbf{W}=\mathbf{B}\mathbf{A}$ inherently consists of~$r$ rank-one components. This provides a natural and meaningful way to turn off unnecessary model capacity. Instead of modeling uncertainty over LLM weights or LoRA adapters, DALorRA leverages this insight to model uncertainty directly at the rank level (Figure \ref{fig:masklora_overview}c). 

Specifically, given a linear layer with a frozen pre-trained weight $\mathbf{W}_0$ and maximum allowable rank $r$, we introduce a stochastic diagonal mask matrix $\mathbf{D}$ and define the weight update as
\begin{equation}
    \Delta \mathbf{W} = \mathbf{B}\mathbf{D}\mathbf{A},
    \label{eq:dalorra_update}
\end{equation}
where $\mathbf{B}\in\mathbb{R}^{d_{\mathrm{out}}\times r}$ and $\mathbf{A}\in\mathbb{R}^{r\times d_{\mathrm{in}}}$ are deterministic adapters, and $\mathbf{D}=\text{diag}(\mathbf{z})$, $\mathbf{z} \in \{0, 1\}^r$, is an $r$-dimensional random diagonal matrix. Each diagonal element $z_j \in \{0, 1\}$ for $j=1,\ldots,r$, dynamically activates or deactivates the $j$-th rank component. 

We assign an independent and identical Bernoulli prior $p(z_j) = \text{Bernoulli}(z_j; p_0)$ over the mask with a constant $p_0 \in (0, 1)$. The exact posterior inference is intractable, and we approximate it with a factorized Bernoulli variational distribution parameterized by $\boldsymbol{\phi} = (\phi_1,\ldots,\phi_r) \in \mathbb{R}^r$, which serves as the  learnable variational parameters: 
\begin{equation*}
    q_{\boldsymbol{\phi}}(\mathbf{z}) = \prod_{j=1}^r \text{Bernoulli}(z_j; \sigma(\phi_j)),
\end{equation*}
where $\sigma(\cdot)$ is the sigmoid function. The deterministic adapters $\mathbf{A}$ and $\mathbf{B}$ and the variational parameters $\boldsymbol{\phi}$ are jointly learned by minimizing the negative Evidence Lower Bound:
\begin{align}
 \mathcal{L} ( \mathbf{A}, \mathbf{B}, \boldsymbol{\phi}) = &- \mathbb{E}_{q_{\boldsymbol{\phi}}(\mathbf{z})} [\log p(\mathcal{D} \mid \mathbf{A}, \mathbf{B}, \mathbf{z})] +\nonumber\\
    &\text{KL}(q_{\boldsymbol{\phi}}(\mathbf{z}) \| p(\mathbf{z})).
    \label{eq:loss1}    
\end{align}

The first term in Eq.~\eqref{eq:loss1} represents the expected negative log-likelihood with weight updates by Eq.~\eqref{eq:dalorra_update}. Since it lacks an analytical solution, we evaluate it via Monte Carlo approximation. To enable gradient-based optimization through the discrete Bernoulli mask, we employ the Gumbel-Sigmoid reparameterization \citep{jang2016categorical}. Specifically, a continuous relaxation of $\mathbf{z} \sim q_{\boldsymbol{\phi}}(\mathbf{z})$ is sampled via
\begin{equation}
    \mathbf{z} = \sigma((\boldsymbol{\phi} + \boldsymbol{\epsilon}) / \tau), \label{eq:gumbel}
\end{equation}
where $\boldsymbol{\epsilon} = \log \mathbf{u} - \log(1 - \mathbf{u})$, $\mathbf{u} \sim \text{Uniform}(0, 1)^r$, and $\tau > 0$ is the temperature parameter. 

The second term in Eq.~\eqref{eq:loss1} is the KL divergence, which regularizes the approximate posterior towards the prior. Because both $p(\mathbf{z})$ and $q_{\boldsymbol{\phi}}(\mathbf{z})$ are Bernoulli distributions, it has the analytical form
$$
\sum_{j=1}^r [ \sigma(\phi_j) \log \frac{\sigma(\phi_j)}{p_0} + (1 - \sigma(\phi_j)) \log \frac{1 - \sigma(\phi_j)}{1 - p_0} ].$$

Substituting the Gumbel-Sigmoid sampling and the analytical KL divergence into Eq.~\eqref{eq:loss1} yields the ultimate differentiable objective function:
\begin{equation}
\begin{aligned}
    \mathcal{L}(\mathbf{A},\mathbf{B},\boldsymbol{\phi}) =& - \frac{1}{S} \sum_{s=1}^{S} \log p(\mathcal{D} \mid \mathbf{A}, \mathbf{B}, \mathbf{z}_{s}) \\
    &+ \sum_{j=1}^{r} \big[ \sigma(\phi_j) \log \frac{\sigma(\phi_j)}{p_0} +\\& (1-\sigma(\phi_j)) \log \frac{1-\sigma(\phi_j)}{1-p_0}\big],
\end{aligned}
\label{eq:finalloss}
\end{equation}
where $\mathbf{z}_s$ for $s=1,\ldots,S$ are independently sampled using Eq.~\eqref{eq:gumbel}. 

To ensure stable optimization and prevent KL divergence explosion, we initialize  $q_{\phiv}(\zv)$ equal to $p_0(\zv)$ by setting $\boldsymbol{\phi} =\log(p_0/(1-p_0)) \cdot \mathbf{1}_r$.  We summarize DALorRA fine-tuning in Algorithm~\ref{alg:dalorra}. 

\begin{algorithm}[!t]
\caption{DALorRA Fine-Tuning}
\label{alg:dalorra}
\textbf{Input:} Dataset $\mathcal{D}$, frozen weights $\mathbf{W}_0$, prior Bernoulli probability $p_0$, Gumbel-sigmoid temperature $\tau$, Monte Carlo sample size $S$ 

\textbf{Output:}  $\mathbf{B}, \mathbf{A}$, and $\boldsymbol{\phi}$
\begin{algorithmic}[1]
\State Initialize $\mathbf{A}, \mathbf{B}$ using standard LoRA initialization and $\boldsymbol{\phi} \leftarrow \log(p_0/(1-p_0)) \cdot \mathbf{1}_r$
\While{not stopped}
\State Sample a mini-batch of $\mathcal{D}$.
\State Sample $\mathbf{z}_1,\ldots,\mathbf{z}_S$  by Eq.~\eqref{eq:gumbel}
\State Compute the mini-batch objective by Eq.~\eqref{eq:finalloss} and update $\mathbf{B}$, $\mathbf{A}$, and $\boldsymbol{\phi}$ by stochastic gradient descent
\EndWhile
\end{algorithmic}
\end{algorithm}

At inference time, the predictive distribution is obtained by averaging over $M$ mask matrices sampled from the learned variational posterior. Specifically, the estimated predictive probability $\hat p(y \mid \mathbf{x}, \mathbf{W}_0, \mathbf{B}, \mathbf{A})
$ is 
\begin{equation*}
     \frac{1}{M} \sum_{m=1}^M p(y \mid \mathbf{x}, \mathbf{W}_0,\mathbf{B}, \mathbf{A}, \mathbf{D}_m),
\end{equation*}
where $\mathbf{D}_m=\text{diag}(\mathbf{z}_m)$ and $\mathbf{z}_m \overset{iid}{\sim} q_{\boldsymbol{\phi}}(\mathbf{z})$. Empirically, setting $M = 10$ \citep{wang2024blob,shi2026training,rahmati2026c} achieves competitive performance with acceptable inference overhead.

Intuitively, the mechanism that allows DALorRA to excel stems from combining the theoretical rigor of Bayesian neural networks with the empirical soundness of deep ensembles, entirely within a parameter-efficient regime. The DALorRA weight update in Eq.~\eqref{eq:dalorra_update} essentially is 
\begin{equation}
    \Delta \mathbf{W} = \sum_{j=1}^r z_j \mathbf{b}_j \mathbf{a}_j,
\end{equation}
where $\mathbf{b}_j$ is the $j$-th column of $\mathbf{B}$, and $\mathbf{a}_j$ is the $j$-th row of $\mathbf{A}$. This effectively trains an ensemble of sub-networks with stochastic weights $z_j$ whose distributions are to be learned. During training, DALorRA acts as a parameter-efficient Bayesian neural network, capturing epistemic uncertainty and penalizing unnecessary complexity through learning the mask posterior distribution. At inference time, averaging predictions over multiple sampled masks mirrors the behavior of a deep ensemble, aggregating outputs across diverse rank-one LoRAs with data-adaptive weights. By pruning superfluous capacity (via some $z_j$ equal to $0$), DALorRA marries the principled uncertainty quantification of Bayesian methods with the robust generalization of ensembles, mitigating overconfidence while retaining the LoRA-like efficiency.

\section{Experiments}
\label{sec:experiments}
In this section, we conduct comprehensive experiments to evaluate the effectiveness of our proposed DALorRA. We first introduce the experimental settings and then report the consistently outstanding performance of DALorRA on LLM calibration without sacrificing reasoning accuracy. We also provide analyses to show the importance of learning the rank mask posterior and the necessity of sufficient model capacity combined with rank sparsity, and illustrate the posterior mask probabilities that vary across transformer layers, projection modules, and data.
More implementation details and experiment results are deferred to Appendices~\ref{app:implementation_details} and~\ref{app:additional_results}.

\begin{table*}[t]
\centering
\setlength{\tabcolsep}{2.8pt}
\caption{
Performance comparison. All methods use Llama3.1-8B pre-trained weights. Accuracy (ACC) and Expected Calibration Error (ECE) are reported in percentages, and Negative Log-Likelihood (NLL) is reported on its original scale.   ``$\uparrow$'' and ``$\downarrow$'' indicate that higher and lower values are preferred, respectively. \textbf{Boldface} and \underline{underlined} numbers denote the best and the second-best performance, respectively.
}
\label{tab:main_results_llama31_8b}
\resizebox{\textwidth}{!}{
\begin{tabular}{llcccccccccc}
\toprule
& \multirow{3}{*}{Method}
& \multicolumn{6}{c}{In-Distribution}
& \multicolumn{4}{c}{Out-of-Distribution (OBQA$\rightarrow$X)} \\
\cmidrule(lr){3-8} \cmidrule(lr){9-12}
& &
& & & & & &
\multicolumn{2}{c}{\textit{Small Shift}}
& \multicolumn{2}{c}{\textit{Large Shift}} \\
\cmidrule(lr){9-10} \cmidrule(lr){11-12}
&
& WG-S & ARC-C & ARC-E & WG-M & OBQA & BoolQ
& ARC-C & ARC-E & Chem & Phy \\
\midrule

\multirow{9}{*}{ACC~($\uparrow$)}

& MCD
& $\secondval{78.03_{\pm0.61}}$ & $81.64_{\pm1.79}$ & $91.37_{\pm0.38}$ & $\secondval{83.18_{\pm0.84}}$ & $87.20_{\pm1.02}$ & $\secondval{89.93_{\pm0.16}}$
& $81.42_{\pm1.38}$ & $\bestval{87.27_{\pm0.84}}$ & $47.92_{\pm2.25}$ & $\secondval{46.53_{\pm0.49}}$ \\

& ENS
& $\bestval{78.82_{\pm0.52}}$ & $\bestval{82.55_{\pm0.42}}$ & $\secondval{91.84_{\pm0.36}}$ & $\bestval{83.99_{\pm0.74}}$ & $87.37_{\pm0.67}$ & $\bestval{90.50_{\pm0.14}}$
& $79.62_{\pm0.57}$ & $86.56_{\pm0.60}$ & $49.65_{\pm3.22}$ & $44.44_{\pm1.96}$ \\

& LA
& $76.05_{\pm0.92}$ & $79.95_{\pm0.42}$ & $90.73_{\pm0.08}$ & $82.83_{\pm0.85}$ & $\secondval{87.90_{\pm0.20}}$ & $89.36_{\pm0.52}$
& $81.08_{\pm1.20}$ & $\secondval{87.21_{\pm1.20}}$ & $48.26_{\pm3.93}$ & $46.18_{\pm1.30}$ \\

& MLE
& $77.87_{\pm0.54}$ & $81.08_{\pm0.48}$ & $91.67_{\pm0.36}$ & $82.30_{\pm0.53}$ & $\secondval{87.90_{\pm0.87}}$ & $89.58_{\pm0.26}$
& $\secondval{81.48_{\pm2.41}}$ & $86.83_{\pm0.87}$ & $45.83_{\pm0.85}$ & $42.36_{\pm1.77}$ \\

& MAP
& $76.90_{\pm0.97}$ & $81.08_{\pm2.48}$ & $91.61_{\pm0.44}$ & $82.59_{\pm0.28}$ & $85.73_{\pm0.19}$ & $90.09_{\pm0.28}$
& $79.98_{\pm0.87}$ & $86.58_{\pm0.79}$ & $43.40_{\pm4.98}$ & $38.54_{\pm3.40}$ \\

& SoRA
& $75.78_{\pm0.79}$ & $79.82_{\pm0.20}$ & $89.57_{\pm0.63}$ & $82.13_{\pm0.72}$ & $86.78_{\pm3.83}$ & $88.65_{\pm4.18}$
& $69.79_{\pm1.51}$ & $79.88_{\pm1.91}$ & $\bestval{50.35_{\pm1.59}}$ & $47.08_{\pm0.02}$ \\

& DropLoRA
& $74.00_{\pm1.05}$ & $77.70_{\pm1.79}$ & $89.73_{\pm0.71}$ & $82.12_{\pm0.70}$ & $87.13_{\pm1.03}$ & $89.41_{\pm0.11}$
& $79.28_{\pm0.40}$ & $85.36_{\pm0.54}$ & $49.65_{\pm4.21}$ & $40.63_{\pm2.76}$ \\

& BLoB
& $77.34_{\pm0.25}$ & $80.86_{\pm1.24}$ & $90.83_{\pm0.68}$ & $81.64_{\pm0.62}$ & $87.66_{\pm0.37}$ & $88.69_{\pm1.26}$
& $78.68_{\pm0.24}$ & $86.63_{\pm0.18}$ & $43.75_{\pm2.65}$ & $46.96_{\pm3.31}$ \\

& TFB
& $74.65_{\pm1.36}$ & $80.18_{\pm1.37}$ & $\bestval{91.90_{\pm0.30}}$ & $82.04_{\pm0.24}$ & $88.20_{\pm0.20}$ & $88.84_{\pm0.21}$
& $81.12_{\pm1.32}$ & $86.81_{\pm0.49}$ & $42.65_{\pm5.12}$ & $46.58_{\pm1.98}$ \\

& C-LoRA 
& $77.26_{\pm0.12}$ & $81.70_{\pm1.17}$ & $90.79_{\pm0.51}$ & $81.62_{\pm0.56}$ & $86.93_{\pm1.62}$ & $87.77_{\pm0.64}$
& $\bestval{81.60_{\pm0.35}}$ & $85.48_{\pm0.55}$ & $45.64_{\pm3.76}$ & $40.38_{\pm3.76}$ \\

\cmidrule(lr){2-12}

\rowcolor{gray!15}
& DALorRA
& $77.43_{\pm1.24}$ & $\secondval{81.74_{\pm0.23}}$ & $90.97_{\pm0.68}$ & $82.68_{\pm0.54}$ & $\bestval{88.24_{\pm0.42}}$ & $89.43_{\pm0.24}$
& $\bestval{81.60_{\pm0.20}}$ & $86.56_{\pm0.20}$ & $\secondval{50.00_{\pm1.08}}$ & $\bestval{47.22_{\pm0.60}}$ \\

\midrule

\multirow{9}{*}{ECE~($\downarrow$)}

& MCD
& $16.13_{\pm0.54}$ & $13.69_{\pm1.11}$ & $6.73_{\pm0.71}$ & $13.05_{\pm0.99}$ & $9.76_{\pm0.71}$ & $7.95_{\pm0.17}$
& $13.63_{\pm1.18}$ & $9.27_{\pm0.60}$ & $30.91_{\pm3.57}$ & $33.08_{\pm1.40}$ \\

& ENS
& $14.72_{\pm0.17}$ & $13.45_{\pm1.19}$ & $6.59_{\pm0.45}$ & $11.17_{\pm0.92}$ & $8.17_{\pm0.86}$ & $7.35_{\pm0.55}$
& $11.37_{\pm1.82}$ & $7.21_{\pm1.13}$ & $18.92_{\pm6.03}$ & $26.80_{\pm3.23}$ \\

& LA
& $\bestval{4.18_{\pm0.11}}$ & $9.26_{\pm3.08}$ & $5.27_{\pm0.51}$ & $3.50_{\pm0.78}$ & $8.93_{\pm0.34}$ & $\secondval{1.93_{\pm0.22}}$
& $7.83_{\pm1.49}$ & $7.80_{\pm1.99}$ & $\secondval{14.49_{\pm0.57}}$ & $\secondval{13.17_{\pm2.14}}$ \\

& MLE
& $17.02_{\pm0.46}$ & $16.35_{\pm0.68}$ & $7.00_{\pm0.53}$ & $13.83_{\pm0.65}$ & $9.77_{\pm0.81}$ & $8.69_{\pm0.21}$
& $14.45_{\pm2.19}$ & $10.78_{\pm0.50}$ & $32.46_{\pm2.60}$ & $38.41_{\pm4.44}$ \\

& MAP
& $18.71_{\pm0.74}$ & $15.77_{\pm1.60}$ & $6.62_{\pm0.64}$ & $14.26_{\pm0.92}$ & $12.19_{\pm0.55}$ & $8.40_{\pm0.25}$
& $16.46_{\pm0.44}$ & $11.36_{\pm0.58}$ & $34.79_{\pm3.76}$ & $38.50_{\pm2.18}$ \\

& SoRA
& $15.91_{\pm0.24}$ & $10.24_{\pm1.10}$ & $6.53_{\pm0.40}$ & $12.87_{\pm0.87}$ & $8.68_{\pm1.40}$ & $5.67_{\pm5.54}$
& $7.42_{\pm2.04}$ & $8.32_{\pm2.65}$ & $\bestval{13.87_{\pm1.43}}$ & $16.19_{\pm2.00}$ \\

& DropLoRA
& $16.07_{\pm0.83}$ & $12.67_{\pm0.90}$ & $9.49_{\pm0.72}$ & $12.99_{\pm0.69}$ & $10.19_{\pm0.66}$ & $4.42_{\pm0.34}$
& $16.41_{\pm1.19}$ & $11.68_{\pm0.72}$ & $30.15_{\pm4.35}$ & $38.27_{\pm4.21}$ \\

& BLoB
& $8.84_{\pm0.36}$ & $\secondval{5.87_{\pm1.12}}$ & $4.24_{\pm0.68}$ & $\secondval{3.42_{\pm0.42}}$ & $\secondval{3.35_{\pm0.82}}$ & $2.46_{\pm0.35}$
& $\secondval{7.02_{\pm0.46}}$ & $5.12_{\pm0.88}$ & $14.79_{\pm0.66}$ & $\bestval{12.34_{\pm3.68}}$ \\

& TFB
& $8.23_{\pm0.68}$ & $6.19_{\pm0.86}$ & $\secondval{3.00_{\pm0.92}}$ & $3.59_{\pm0.64}$ & $4.51_{\pm0.33}$ & $3.80_{\pm0.61}$
& $7.12_{\pm1.22}$ & $\secondval{4.85_{\pm0.36}}$ & $15.32_{\pm2.12}$ & $16.32_{\pm4.37}$ \\

& C-LoRA 
& $16.84_{\pm1.88}$ & $10.75_{\pm0.81}$ & $4.95_{\pm1.43}$ & $9.97_{\pm1.02}$ & $6.50_{\pm1.52}$ & $4.36_{\pm0.90}$
& $7.06_{\pm0.83}$ & $5.63_{\pm0.94}$ & $26.48_{\pm2.19}$ & $30.28_{\pm4.12}$ \\

\cmidrule(lr){2-12}

\rowcolor{gray!15}
& DALorRA
& $\secondval{7.81_{\pm1.08}}$ & $\bestval{5.60_{\pm0.64}}$ & $\bestval{2.88_{\pm0.37}}$ & $\bestval{3.25_{\pm0.39}}$ & $\bestval{3.12_{\pm0.49}}$ & $\bestval{1.82_{\pm0.22}}$
& $\bestval{6.23_{\pm0.79}}$ & $\bestval{3.81_{\pm1.09}}$ & $14.58_{\pm1.65}$ & $15.46_{\pm3.08}$ \\

\midrule

\multirow{9}{*}{NLL~($\downarrow$)}

& MCD
& $0.83_{\pm0.01}$ & $0.99_{\pm0.10}$ & $0.45_{\pm0.06}$ & $0.64_{\pm0.03}$ & $0.62_{\pm0.08}$ & $0.49_{\pm0.01}$
& $1.03_{\pm0.02}$ & $0.61_{\pm0.03}$ & $1.91_{\pm0.18}$ & $2.02_{\pm0.15}$ \\

& ENS
& $0.75_{\pm0.02}$ & $0.80_{\pm0.11}$ & $0.38_{\pm0.03}$ & $0.55_{\pm0.02}$ & $0.45_{\pm0.05}$ & $0.42_{\pm0.05}$
& $0.72_{\pm0.07}$ & $\secondval{0.44_{\pm0.03}}$ & $1.40_{\pm0.18}$ & $1.50_{\pm0.13}$ \\

& LA
& $\bestval{0.56_{\pm0.00}}$ & $1.18_{\pm0.02}$ & $1.04_{\pm0.01}$ & $0.51_{\pm0.00}$ & $0.94_{\pm0.00}$ & $0.43_{\pm0.00}$
& $1.17_{\pm0.01}$ & $1.11_{\pm0.00}$ & $1.27_{\pm0.01}$ & $1.28_{\pm0.00}$ \\

& MLE
& $0.88_{\pm0.04}$ & $1.20_{\pm0.11}$ & $0.46_{\pm0.04}$ & $0.68_{\pm0.01}$ & $0.61_{\pm0.06}$ & $0.52_{\pm0.01}$
& $1.07_{\pm0.06}$ & $0.72_{\pm0.06}$ & $1.91_{\pm0.16}$ & $2.25_{\pm0.21}$ \\

& MAP
& $0.99_{\pm0.07}$ & $1.12_{\pm0.23}$ & $0.46_{\pm0.03}$ & $0.74_{\pm0.07}$ & $0.79_{\pm0.02}$ & $0.52_{\pm0.01}$
& $1.19_{\pm0.04}$ & $0.83_{\pm0.06}$ & $1.97_{\pm0.13}$ & $2.32_{\pm0.10}$ \\

& SoRA
& $0.92_{\pm0.15}$ & $0.82_{\pm0.11}$ & $0.50_{\pm0.04}$ & $0.70_{\pm0.01}$ & $0.82_{\pm0.08}$ & $0.49_{\pm0.03}$
& $0.79_{\pm0.05}$ & $0.58_{\pm0.05}$ & $\bestval{1.16_{\pm0.02}}$ & $1.31_{\pm0.01}$ \\

& DropLoRA
& $1.26_{\pm0.18}$ & $1.13_{\pm0.42}$ & $0.81_{\pm0.03}$ & $0.93_{\pm0.01}$ & $0.62_{\pm0.06}$ & $0.29_{\pm0.01}$
& $1.03_{\pm0.03}$ & $0.80_{\pm0.04}$ & $1.98_{\pm0.20}$ & $2.12_{\pm0.28}$ \\

& BLoB
& $0.58_{\pm0.01}$ & $\secondval{0.59_{\pm0.02}}$ & $0.30_{\pm0.01}$ & $\secondval{0.47_{\pm0.01}}$ & $0.38_{\pm0.01}$ & $\bestval{0.27_{\pm0.01}}$
& $0.61_{\pm0.03}$ & $0.46_{\pm0.01}$ & $\secondval{1.23_{\pm0.06}}$ & $1.28_{\pm0.22}$ \\

& TFB
& $0.59_{\pm0.01}$ & $0.62_{\pm0.04}$ & $\bestval{0.25_{\pm0.01}}$ & $\bestval{0.43_{\pm0.01}}$ & $\bestval{0.36_{\pm0.02}}$ & $\secondval{0.29_{\pm0.02}}$
& $0.60_{\pm0.04}$ & $\secondval{0.44_{\pm0.02}}$ & $1.34_{\pm0.08}$ & $\secondval{1.26_{\pm0.12}}$ \\

& C-LoRA
& $0.99_{\pm0.11}$ & $0.68_{\pm0.03}$ & $0.33_{\pm0.06}$ & $0.57_{\pm0.06}$ & $0.40_{\pm0.04}$ & $0.30_{\pm0.01}$
& $\secondval{0.55_{\pm0.03}}$ & $0.48_{\pm0.03}$ & $1.66_{\pm0.16}$ & $1.88_{\pm0.09}$ \\
\cmidrule(lr){2-12}
\rowcolor{gray!15}
& DALorRA
& $\secondval{0.58_{\pm0.01}}$ & $\bestval{0.56_{\pm0.02}}$ & $\secondval{0.28_{\pm0.01}}$ & $0.51_{\pm0.01}$ & $\bestval{0.36_{\pm0.02}}$ & $\bestval{0.27_{\pm0.01}}$
& $\bestval{0.54_{\pm0.07}}$ & $\bestval{0.40_{\pm0.01}}$ & $1.25_{\pm0.02}$ & $\bestval{1.25_{\pm0.06}}$ \\

\bottomrule
\end{tabular}
}

\end{table*}
\subsection{Experiment Settings}
\label{sec:experimental_settings}
\paragraph{LLM fine-tuning.} 
We fine-tune Llama-3.1-8B \citep{grattafiori2024llama} and Llama2-7B \citep{touvron2023llama} using the PEFT library \citep{mangrulkar2022peft} on common-sense reasoning benchmarks, including Winogrande-Small (WG-S) \citep{sakaguchi2021winogrande}, Winogrande-Medium (WG-M) \citep{sakaguchi2021winogrande}, ARC-Challenge (ARC-C) \citep{clark2018think}, ARC-Easy (ARC-E) \citep{clark2018think}, OpenBookQA (OBQA) \citep{mihaylov2018can}, and BoolQ \citep{clark2019boolq}. For out-of-distribution evaluation, we fine-tune models on OBQA \citep{mihaylov2018can} and evaluate on ARC-C \citep{clark2018think}, ARC-E \citep{clark2018think}, and college-level chemistry ({Chem}) and physics ({Phy})~\citep{hendrycks2020measuring}. 

\paragraph{Evaluation metrics.} \looseness-1
We report accuracy (ACC), expected calibration error (ECE)~\citep{naeini2015obtaining}, and negative log-likelihood (NLL). 
ACC measures predictive performance. ECE and NLL evaluate calibration and probabilistic reliability. 
Higher ACC and lower ECE and NLL are preferred. 

\paragraph{Baseline methods.}
We compare DALorRA with representative deterministic and uncertainty-aware baselines for LLM fine-tuning. The deterministic baselines include Maximum Likelihood Estimation (MLE) implemented with standard LoRA \citep{hu2022lora}, and maximum a posteriori(MAP) estimation , which serves as a point-estimate baseline with weight-decay regularization. For uncertainty-aware baselines, we include Monte Carlo Dropout (MCD) \citep{gal2016dropout}, Deep Ensemble (ENS) \citep{lakshminarayanan2017simple,balabanov2024uncertainty,wang2023lora}, Laplace-LoRA (LA) \citep{yang2024bayesian}, Bayesian LoRA by Backpropagation (BLoB) \citep{wang2024blob}, Training-Free Bayesianization (TFB) \citep{shi2026training}, and Contextual Low-Rank Adaptation (C-LoRA) \citep{rahmati2026c}.
We use $M=10$ Monte Carlo samples for inference \citep{shi2026training} in all sampling-based methods, including BLoB, TFB, C-LoRA, and DALorRA.

\begin{table*}[t]
\centering
\renewcommand{\arraystretch}{1.22}
\setlength{\tabcolsep}{3.0pt}
\caption{{Efficiency Comparison among LoRA, BLoB, C-LoRA, and DALorRA.} The experiments are conducted on a single NVIDIA A40 GPU and based on Llama-3.1-8B with rank $r=8$, batch size 4, and 2,000 training iterations. The subscripts indicate the relative cost compared to LoRA, with \textcolor{red}{red} or \textcolor{green!50!black}{green} denoting worse or better efficiency.}
\label{tab:resource_consumption_all}
\resizebox{\textwidth}{!}{
\begin{tabular}{l c cc cc cc}
\toprule
\multirow{2}{*}{\textbf{Method}}
& \multirow{2}{*}{\textbf{Trainable/Extra Params}}
& \multicolumn{2}{c}{\textbf{WG-S}}
& \multicolumn{2}{c}{\textbf{ARC-E}}
& \multicolumn{2}{c}{\textbf{OBQA}} \\
\cmidrule(lr){3-4} \cmidrule(lr){5-6} \cmidrule(lr){7-8}
&
& \textbf{Train/Eval (s)} & \textbf{Mem. (MB)}
& \textbf{Train/Eval (s)} & \textbf{Mem. (MB)}
& \textbf{Train/Eval (s)} & \textbf{Mem. (MB)} \\
\midrule

LoRA
& $4{,}466{,}688 / 0$
& $1{,}237.07 / 104.19$
& $7{,}690.08$
& $1{,}799.69 / 66.76$
& $11{,}296.05$
& $1{,}524.87 / 50.99$
& $9{,}410.75$ \\

BLoB
& $6{,}596{,}608_{\textcolor{red}{1.48\times}} / +2{,}129{,}920$
& $1{,}316.11_{\textcolor{red}{1.06\times}} / 1{,}058.12_{\textcolor{red}{10.16\times}}$
& $8{,}214.50_{\textcolor{red}{1.07\times}}$
& $1{,}901.34_{\textcolor{red}{1.06\times}} / 683.49_{\textcolor{red}{10.24\times}}$
& $12{,}874.76_{\textcolor{red}{1.14\times}}$
& $1{,}610.29_{\textcolor{red}{1.06\times}} / 520.53_{\textcolor{red}{10.21\times}}$
& $10{,}453.27_{\textcolor{red}{1.11\times}}$ \\

C-LoRA
& $5{,}044{,}928_{\textcolor{red}{1.13\times}} / +578{,}240$
& $4{,}640.89_{\textcolor{red}{3.75\times}} / 714.55_{\textcolor{red}{6.86\times}}$
& $11{,}245.92_{\textcolor{red}{1.46\times}}$
& $4{,}911.09_{\textcolor{red}{2.73\times}} / 366.43_{\textcolor{red}{5.49\times}}$
& $15{,}718.29_{\textcolor{red}{1.39\times}}$
& $4{,}745.12_{\textcolor{red}{3.11\times}} / 340.10_{\textcolor{red}{6.67\times}}$
& $13{,}410.96_{\textcolor{red}{1.43\times}}$ \\

DALorRA (Ours)
& $4{,}467{,}208_{\textcolor{red}{1.0001\times}} / +520$
& $894.31_{\textcolor{green!50!black}{0.72\times}} / 809.60_{\textcolor{red}{7.77\times}}$
& $9{,}133.99_{\textcolor{red}{1.19\times}}$
& $1{,}078.27_{\textcolor{green!50!black}{0.60\times}} / 414.98_{\textcolor{red}{6.22\times}}$
& $13{,}670.46_{\textcolor{red}{1.21\times}}$
& $990.34_{\textcolor{green!50!black}{0.65\times}} / 346.64_{\textcolor{red}{6.80\times}}$
& $11{,}336.79_{\textcolor{red}{1.20\times}}$ \\

\bottomrule
\end{tabular}
}
\end{table*}

\subsection{Main Results}\label{sec:mainresult}

\begin{figure*}[t]
    \centering
    \includegraphics[width=0.9\textwidth]{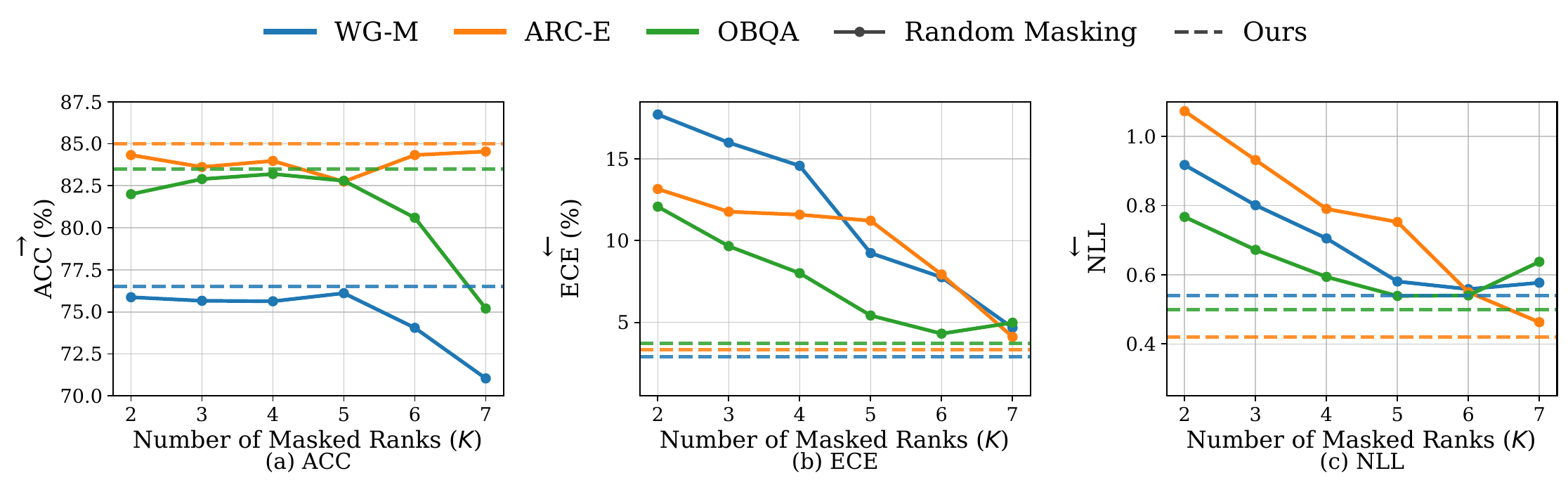}
    \caption{
    Learning mask posterior (DALorRA) versus random masking. 
    Solid lines denote randomly dropping out  $K$ rank dimensions during training, and dashed lines denote DALorRA.
    }
    \label{fig:rank_mask_selection}
\end{figure*}
\paragraph{Calibration performance.}
Table~\ref{tab:main_results_llama31_8b} reports the in- and out-of-distribution performance of different LoRA-based uncertainty estimation methods on Llama3.1-8B. 
Overall, Table~\ref{tab:main_results_llama31_8b} demonstrates DALorRA's exceptional uncertainty calibration and reliability without sacrificing reasoning accuracy across both in-distribution and out-of-distribution (OOD) evaluations. On average, DALorRA achieves the strongest performance in both the reasoning (ACC) and calibration (ECE and NLL) metrics across the 10 evaluation settings. Specifically, when evaluated by ECE and NLL, DALorRA consistently outperforms the baselines, achieving the best or second-best performance in 9 of the 10 settings for both ECE and NLL. Moreover, it achieves this without compromising reasoning capabilities (ACC). In particular, DALorRA maintains highly competitive predictive performance compared to the baseline approaches, achieving the best or second-best ACC on 5 out of the 10 settings. The results suggest that DALorRA's uncertainty over rank remarkably mitigates overconfidence. 

On in-distribution datasets, DALorRA achieves the best ECE on five benchmarks and top-tier NLL, offering consistently better uncertainty estimates. The reliability extends to OOD evaluations, where DALorRA secures the best ECE and NLL under small shifts, and the highest accuracy alongside highly competitive NLL under large shifts (Chem and Phy). Ultimately, these results confirm that learning a structured posterior over LoRA rank effectively captures uncertainty without sacrificing reasoning capabilities, despite distribution shifts.

\paragraph{Efficiency, resource consumption, and deterministic inference via weight scaling.}\label{sec:resource_consumption} 
We compare the computational efficiency of the sampling-based uncertainty-aware fine-tuning methods (BLoB, C-LoRA, and DALorRA) with LoRA.
Table~\ref{tab:resource_consumption_all} shows DALorRA's high efficiency. It adds only 520 extra trainable parameters, which is negligible and much fewer than BLoB and C-LoRA. 
DALorRA takes the least training time, even faster than LoRA due to its sparse configuration; it uses only 0.72×, 0.60×, and 0.65× of LoRA's training time on WG-S, ARC-E, and OBQA, respectively.
DALorRA remains faster than BLoB and comparable to C-LoRA in evaluation (inference) time with moderate memory overhead. 
In addition, DALorRA's overall (training plus evaluation) time is comparable to LoRA's on the three datasets.

We additionally use a deterministic inference for DALorRA analogous to test-time weight scaling in dropout \citep{srivastava2014dropout}; instead of repeatedly sampling Bernoulli rank masks and averaging predictions, we replace mask $\zv$ with its learned posterior mean $\pv$. That is, multiple sampled $\mathbf{B}\operatorname{diag}(\mathbf{z})\mathbf{A}$ is replaced by the deterministic $\mathbf{B}\operatorname{diag}(\mathbf{p})\mathbf{A}$ for inference. The inference time is close to that of LoRA.  To compare the performance, we use the same strategy on Bayesian LoRA baselines, including BLoB, C-LoRA, and TFB, where the parameters to be sampled are replaced by their posterior mean during inference.  Table~\ref{tab:mean_inference_id} in the appendix shows that without MC sampling for inference, DALorRA still achieves competitive performance in uncertainty quantification.

\subsection{Ablation and Additional Analyses}

\label{sec:ablation}
\begin{figure*}[t]
    \centering
    \includegraphics[width=0.9\textwidth]{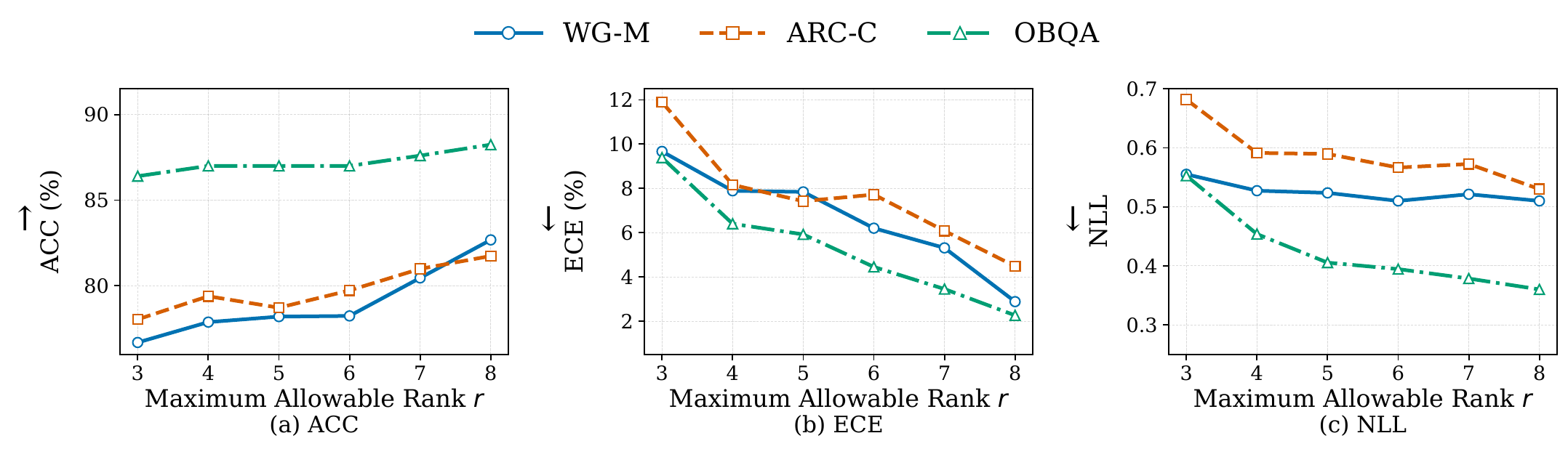}
    \caption{
    Impact of maximum allowable rank $r$. A large $r$ generally improves both accuracy and calibration. }
    \label{fig:rank_capacity}
\end{figure*}
\paragraph{{Necessity of mask posterior learning.}}
To demonstrate the critical need for learning the mask posterior, we compare DALorRA against a random masking strategy that similarly induces rank uncertainty. 
Specifically, we fix $r=8$ and randomly zero out $K$ (up to $7$) diagonal entries of $\mathbf{D}$ in every mini-batch iteration of stochastic gradient descent, acting as a rank-level dropout. 
The inference is the same as in DALorRA, where $M=10$ Monte Carlo samples of $\mathbf{D}$ are aggregated for prediction.

The results are illustrated in Figure~\ref{fig:rank_mask_selection}. Random masking, even with an optimal $K$, still underperforms DALorRA, and $K$ remarkably affects its performance and depends on downstream tasks.
For instance, aggressively masking rank dimensions severely degrades accuracy on WG-M and OBQA due to capacity starvation but slightly improves accuracy on ARC-E. 
Moreover, on WG-M and OBQA, better ECE and NLL are achieved at a smaller $K$, which, however, severely compromises ACC, whereas all three metrics are optimized at a larger $K$ on {ARC-E}. 
This inconsistent accuracy-calibration trade-off highlights the necessity of data-adaptive rank uncertainty. 
By learning the mask posterior and performing posterior sampling for inference, DALorRA achieves consistently strong performance without manually specifying masking intensity.

\paragraph{{Maximum allowable rank.}}
We show the important role of the maximum allowable rank $r$.  
We vary $r$ from $3$ to $8$ to obtain different levels of model capacity and compare
DALorRA’s performance. 
As shown in Figure~\ref{fig:rank_capacity}, 
increasing $r$ generally improves accuracy and reduces ECE and NLL.
This indicates that better performance stems from combining sufficient capacity with rank sparsity. In addition, the performance trends vary across data. For example, in Figure~\ref{fig:rank_capacity}c, DALorRA's NLL decreases with $r$ faster on ARC-C and OBQA than on WG-M, which also necessitates learning the mask posterior from the data. 

\paragraph{Posterior mask probabilities.}\label{sec:posterior_mask_analysis}
Figure~\ref{fig:rank_mask_heatmap} illustrates the learned posterior mask probabilities for the transformer Query and Value projection matrices. We observe non-uniform patterns across both transformer layers and rank indices, indicating that DALorRA's rank sparsity depends on the layer, projection module, and data. 
This qualitative evidence suggests that the proposed masking mechanism learns data-adaptive rank allocation instead of imposing a uniform rank structure. 

\paragraph{Open-Ended Question Answering.}
To assess DALorRA beyond classification-style reasoning, we further evaluate it on open-ended generation using TriviaQA. We quantify predictive confidence by the normalized exact-match agreement among responses generated from multiple posterior samples. Under the same experimental protocol as the main results, DALorRA achieves the best accuracy and AUROC, and the second-best ECE, compared with competitive baselines (Table~\ref{tab:triviaqa}). These results suggest that the learned rank-mask posterior also provides effective uncertainty estimates for open-ended question answering.

\begin{table}[t]
\centering
\caption{Performance comparison on TriviaQA under free-form generation with
Llama-3.1-8B. Accuracy, ECE, and AUROC are reported in percentages.}
\label{tab:triviaqa}
\scriptsize
\setlength{\tabcolsep}{2.5pt}
\begin{tabular}{lccc}
\toprule
\textbf{Method}
& \textbf{Accuracy} $\uparrow$
& \textbf{ECE} $\downarrow$
& \textbf{AUROC} $\uparrow$ \\
\midrule
C-LoRA
& $62.58 \pm 0.04$
& $10.11 \pm 0.04$
& $78.43 \pm 0.24$ \\

BLoB
& $\underline{64.36 \pm 0.88}$
& $\mathbf{4.53 \pm 0.13}$
& $\underline{81.91 \pm 1.47}$ \\

TFB
& $63.54 \pm 0.26$
& $5.48 \pm 0.18$
& $80.66 \pm 1.06$ \\

DALorRA
& $\mathbf{66.88 \pm 0.60}$
& $\underline{5.15 \pm 0.33}$
& $\mathbf{82.10 \pm 0.85}$ \\
\bottomrule
\end{tabular}
\end{table}

\begin{figure}[!t]
    \centering
    \includegraphics[width=0.9\columnwidth]{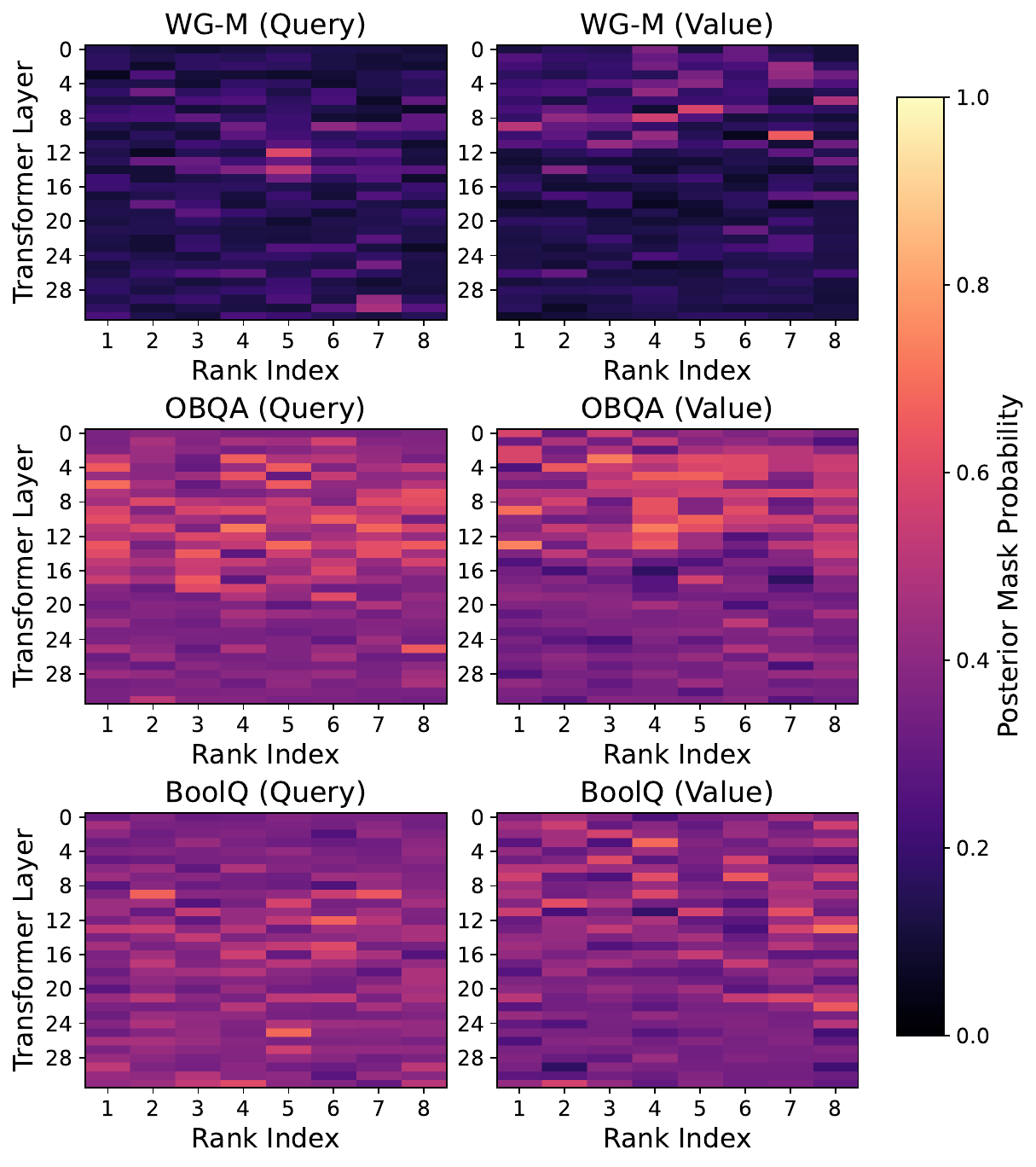}
    \caption{
    Posterior Bernoulli probabilities of DALorRA masks $\mathbf{D}$ on WG-M, OBQA, and BoolQ.
    }
    \label{fig:rank_mask_heatmap}
\end{figure}

\paragraph{Additional results.} We provide additional experimental results in Appendix~\ref{app:additional_results}, including using an alternative foundation model, Llama2-7B, performance on combined datasets, impacts of the Bernoulli prior, and efficiency comparison on other data.
\section{Conclusion}
We propose DALorRA as a novel parameter-efficient fine-tuning framework for LLM uncertainty quantification. By systematically learning a variational posterior over the LoRA rank components, DALorRA prunes superfluous capacity while effectively capturing epistemic uncertainty. Our comprehensive evaluations across diverse tasks demonstrate that DALorRA significantly improves LLM calibration without sacrificing fundamental reasoning accuracy. By introducing negligible parameter overhead and training costs compared to standard LoRA, our approach offers a scalable solution. Bridging the principled uncertainty estimation of Bayesian methods with the robust generalization of deep ensembles, DALorRA provides a practical and reliable pathway for deploying trustworthy LLMs in complex, real-world environments.
\clearpage
\section*{Acknowledgements}
The authors would like to thank the anonymous reviewers for their valuable comments. In this work, Jijie Zhang, Zhe Ren and Dandan Guo are supported by the National Natural Science Foundation of China (No. 62306125).

\section*{Limitations}
While DALorRA presents a highly parameter-efficient approach to LLM calibration, several limitations highlight promising directions for future research. First, our empirical evaluation is predominantly focused on classification-style reasoning tasks with fixed answer spaces; future work can extend this framework to open-ended generation scenarios where hallucinations and uncertainty are more complex to characterize. Second, the reliance on multiple Bernoulli mask samples for inference introduces extra overhead, which poses challenges for latency-sensitive applications and necessitates the development of more efficient inference approximations. Third, because DALorRA models uncertainty solely at the LoRA rank level rather than over full adapter weights, it may miss fine-grained, weight-level uncertainty. To address this, future work could seamlessly integrate our data-adaptive rank masking with more expressive, weight-level Bayesianization approaches. Finally, the current variational posterior assumes an independent Bernoulli factorization that ignores potential dependencies across ranks, layers, and projection modules. Developing structured variational posteriors capable of capturing these complex architectural dependencies remains an important area for future investigation.

\section*{Ethical Considerations}
We anticipate no ethical concerns related to this work, as the study exclusively uses publicly available data with sensitive information redacted.
We used AI writing assistants only for language polishing, formatting suggestions, and auxiliary writing support. All scientific claims, experimental results, analyses, and final manuscript content were verified and approved by the authors.

\bibliography{references}
\clearpage
\appendix

\section{Implementation Details}
\label{app:implementation_details}

\subsection{Dataset Details}
\label{app:dataset_details}

We provide dataset statistics in Table~\ref{tab:dataset_statistics}. 
AAO denotes the merged dataset of ARC-E, ARC-C, and OBQA, while WB denotes the merged dataset of Winogrande-Medium and BoolQ. 
The full Combined dataset merges all six commonsense reasoning benchmarks and uses the unified label set 
$\{\mathrm{A}, \mathrm{B}, \mathrm{C}, \mathrm{D}, \mathrm{E}, \mathrm{True}, \mathrm{False}\}$.
Table~\ref{tab:prompt_templates} summarizes the prompt templates used for commonsense reasoning tasks.

\begin{table*}[t]
\centering
\caption{Dataset statistics.}
\label{tab:dataset_statistics}
\resizebox{\textwidth}{!}{
\begin{tabular}{lccccccccccc}
\toprule
 & WG-S & ARC-C & ARC-E & WG-M & OBQA & BoolQ & AAO & WB & Chem & Phy & Combined \\
\midrule
Size of Label Space 
& 2 & 5 & 5 & 2 & 4 & 2 & 5 & 4 & 4 & 4 & 7 \\
\midrule
Size of Training Set 
& 640 & 1,119 & 2,251 & 2,258 & 4,957 & 9,427 & 8,327 & 11,685 & -- & -- & 20,652 \\
\midrule
Size of Test Set 
& 1,267 & 299 & 570 & 1,267 & 500 & 3,270 & 1,369 & 4,537 & 100 & 102 & 7,173 \\
\bottomrule
\end{tabular}
}
\end{table*}

\begin{table*}[t]
\centering
\caption{Prompt templates for common sense reasoning tasks.}
\label{tab:prompt_templates}
\resizebox{1\textwidth}{!}{
\begin{tabular}{lp{0.68\textwidth}}
\toprule
{Task} & {Prompt} \\
\midrule
Winogrande (WG-S/WG-M) 
& Select one of the choices that answers the following question: \newline
\{question\} Choices: A. \{option1\}. B. \{option2\}. Answer: \\

\midrule
ARC (ARC-C/ARC-E), OpenBookQA (OBQA), MMLU 
& Select one of the choices that answers the following question: \newline
\{question\} Choices: A. \{choice1\}. B. \{choice2\}. C. \{choice3\}. D. \{choice4\}. Answer: \\

\midrule
BoolQ 
& Answer the question with only True or False: \newline
\{question\} Context: \{passage\}. \\

\bottomrule
\end{tabular}
}
\end{table*}

\subsection{Evaluation Metrics for Uncertainty Estimation}
\label{app:evaluation_metrics}

We evaluate predictive performance and uncertainty quality using Accuracy (ACC), Negative Log-Likelihood (NLL), and Expected Calibration Error (ECE)~\citep{naeini2015obtaining}. 
Given a test set $\mathcal{D}_{\mathrm{test}}=\{(\mathbf{x}_i,y_i)\}_{i=1}^{N}$, the predicted label is defined as
$\hat{y}_i=\arg\max_y p_{\theta}(y\mid \mathbf{x}_i)$. 
Accuracy measures the fraction of correctly predicted examples:
\begin{equation}
    \mathrm{ACC}
    =
    \frac{1}{N}
    \sum_{i=1}^{N}
    \mathbf{1}(\hat{y}_i = y_i).
\end{equation}

NLL evaluates the quality of the full predictive distribution by penalizing low probability assigned to the ground-truth label:
\begin{equation}
    \mathrm{NLL}
    =
    -\frac{1}{N}
    \sum_{i=1}^{N}
    \log p_{\theta}(y_i\mid \mathbf{x}_i).
\end{equation}
Compared with accuracy, NLL is more sensitive to over-confident wrong predictions, since assigning high confidence to an incorrect label increases the loss substantially.

ECE measures how well model confidence matches empirical correctness. 
Following standard practice, we partition predictions into $N_{\mathrm{bin}}$ confidence bins $\{B_b\}_{b=1}^{N_{\mathrm{bin}}}$ and compute the weighted average gap between accuracy and confidence:
\begin{equation}
    \mathrm{ECE}
    =
    \sum_{b=1}^{N_{\mathrm{bin}}}
    \frac{|B_b|}{N}
    \left|
    \mathrm{acc}(B_b) - \mathrm{conf}(B_b)
    \right|.
\end{equation}
Here, the accuracy and confidence of bin $B_b$ are computed as
\begin{equation}
    \mathrm{acc}(B_b)
    =
    \frac{1}{|B_b|}
    \sum_{i\in B_b}
    \mathbf{1}(\hat{y}_i = y_i),
\end{equation}
\begin{equation}
    \mathrm{conf}(B_b)
    =
    \frac{1}{|B_b|}
    \sum_{i\in B_b}
    \max_y p_{\theta}(y\mid \mathbf{x}_i).
\end{equation}
In our experiments, we use $N_{\mathrm{bin}}=15$ bins for ECE computation. 
Higher ACC indicates better predictive performance, while lower NLL and ECE indicate better uncertainty estimation and calibration.

\subsection{Bayesianization and Training}
\label{app:bayesianization_training}

\paragraph{Shared Configuration.}
We use the same data splits, prompt templates, and evaluation protocol for all methods. 
Unless otherwise specified, all LoRA-based models are fine-tuned with AdamW using a batch size of 4, a linear learning-rate decay schedule, and a warmup ratio of 0.06. 
The maximum sequence length is set to 300 tokens. 
For standard LoRA adapters, we set the LoRA rank to $r=8$ and the LoRA scaling factor to $\alpha=16$. 
For sampling-based uncertainty estimation methods, we use $M=10$ Monte Carlo samples during inference unless otherwise stated. 
All reported results are averaged over three random seeds when repeated runs are available.

\paragraph{Baseline Configuration.}
We compare DALorRA with representative deterministic and uncertainty-aware LoRA baselines. 
MLE and MAP are used as deterministic PEFT baselines, where MLE follows standard LoRA fine-tuning and MAP introduces prior-based regularization. 
MCD and Deep Ensemble are included as sampling- and ensemble-based baselines: MCD averages multiple predictions with dropout enabled at inference time, while Deep Ensemble averages independently fine-tuned LoRA models. 
LA is implemented as a post-hoc Bayesian method that applies a Laplace approximation to trained LoRA parameters.

For variational Bayesian baselines, we include BLoB, TFB, and C-LoRA. 
BLoB Bayesianizes the LoRA adapter by placing a Gaussian variational posterior over the $\mathbf{A}$ matrix while keeping $\mathbf{B}$ deterministic~\citep{wang2024blob}. 
TFB is applied as a training-free Bayesianization method that converts trained LoRA adapters into Bayesian ones without additional gradient-based fine-tuning~\citep{shi2026training}. 
C-LoRA follows its original contextual low-rank adaptation setting, where a lightweight contextual module is used to model input-dependent uncertainty in the low-rank space~\citep{rahmati2026c}.  All baselines are implemented following the protocols established in TFB~\citep{shi2026training}. 
\paragraph{DALorRA Configuration.}
For DALorRA, we set the maximum allowable rank $r=8$, scaling factor $\alpha=16$, and dropout rate $0$. 
All models are trained for 5,000 gradient steps with AdamW, learning rate $1\times 10^{-4}$, batch size 4, warmup ratio 0.06, and maximum sequence length 300. 
For variational rank masking, we use a separate mask-logit learning rate of $1\times 10^{-2}$.
At inference time, we use $M=10$ Bernoulli mask samples for Bayesian model averaging. 
All results are evaluated under the same protocol as the baseline methods.
\section{Additional Experimental Results}
\label{app:additional_results}

\subsection{Performance on Llama2-7B}
\label{app:llama2_results}
Table~\ref{tab:main_results_llama2_7b} reports additional results on Llama2-7B across six common-sense reasoning datasets. 
Compared with deterministic and ensemble-based baselines, DALorRA maintains competitive accuracy while achieving strong calibration performance. 
In particular, DALorRA obtains the best ECE on five out of six datasets and the second-best ECE on ARC-E, showing consistent improvements in confidence calibration. 
For NLL, DALorRA also achieves the best or competitive performance on most datasets, especially on ARC-C, WG-M, and OBQA. 
These results further confirm that posterior rank masking can improve uncertainty estimation without causing severe degradation in predictive accuracy.
\begin{table*}[t]
\centering
\caption{{Performance of different methods applied to LoRA on Llama2-7B pre-trained weights}, where Accuracy ({ACC}) and Expected Calibration Error ({ECE}) are reported in percentages. The evaluation is done across six common-sense reasoning tasks with a shared hyperparameter setting after 5,000 gradient steps. We use $M = 10$ samples during inference in all sampling-based methods, including {BLoB}~\citep{wang2024blob}, {TFB}~\citep{shi2026training}, and {C-LoRA}~\citep{rahmati2026c}. ``$\uparrow$'' and ``$\downarrow$'' indicate that higher and lower values are preferred, respectively. \textbf{Boldface} and \underline{underlining} denote the best and the second-best performance, respectively.}
\label{tab:main_results_llama2_7b}
\resizebox{\textwidth}{!}{
\begin{tabular}{llcccccc}
\toprule
\multirow{2}{*}{{Metric}} & \multirow{2}{*}{{Method}} & \multicolumn{6}{c}{{Data}} \\
\cmidrule(lr){3-8}
& & WG-S  & ARC-C & ARC-E  & WG-M  & OBQA & BoolQ  \\
\midrule

\multirow{8}{*}{ACC (\uparrowcenter)} 
& MAP               & $\best{69.37_{\pm1.04}}$ & $67.67_{\pm1.18}$ & $85.20_{\pm0.63}$ & $74.57_{\pm0.73}$ & $81.60_{\pm0.40}$ & $\second{87.68_{\pm0.02}}$ \\
& MCD               & $\second{69.06_{\pm1.40}}$ & $66.66_{\pm2.30}$ & $\second{85.49_{\pm0.74}}$ & $\second{75.89_{\pm0.48}}$ & $81.46_{\pm0.92}$ & $87.67_{\pm0.08}$ \\
& Deep Ensemble     & $68.98_{\pm0.97}$ & $\best{68.57_{\pm2.11}}$ & $\best{86.24_{\pm1.26}}$ & $\best{77.39_{\pm1.08}}$ & $\best{82.20_{\pm0.91}}$ & $\best{88.07_{\pm0.17}}$ \\
& LA                & $68.18_{\pm1.04}$ & $64.17_{\pm0.97}$ & $85.30_{\pm0.97}$ & $74.15_{\pm0.40}$ & $77.53_{\pm0.80}$ & $86.45_{\pm0.35}$ \\
& BLoB              & $66.55_{\pm0.61}$ & $66.66_{\pm2.25}$ & $84.56_{\pm0.20}$ & $73.38_{\pm0.29}$ & $81.44_{\pm0.53}$ & $86.63_{\pm0.50}$ \\
& TFB               & $66.84_{\pm1.52}$ & $67.62_{\pm1.12}$ & $84.52_{\pm0.62}$ & $73.13_{\pm2.38}$ & $81.10_{\pm0.61}$ & $86.36_{\pm0.26}$ \\
& C-LoRA            & $66.21_{\pm1.24}$ & $\second{67.79_{\pm1.27}}$ & $84.38_{\pm0.67}$ & $70.48_{\pm1.71}$ & $78.26_{\pm2.61}$ & $84.64_{\pm0.81}$ \\
\cmidrule(lr){2-8}
\rowcolor{gray!15}
& DALorRA           & $66.61_{\pm0.87}$ & $67.57_{\pm0.96}$ & $85.04_{\pm0.52}$ & $73.73_{\pm0.75}$ & $\second{81.80_{\pm0.43}}$ & $86.44_{\pm0.28}$ \\
\midrule

\multirow{8}{*}{ECE (\downarrowcenter)} 
& MAP               & $29.76_{\pm1.08}$ & $30.60_{\pm1.26}$ & $13.49_{\pm0.63}$ & $23.01_{\pm0.44}$ & $15.30_{\pm0.11}$ & $5.93_{\pm0.36}$ \\
& MCD               & $28.49_{\pm1.60}$ & $29.60_{\pm2.77}$ & $12.69_{\pm0.60}$ & $20.73_{\pm0.38}$ & $14.34_{\pm1.11}$ & $5.13_{\pm0.25}$ \\
& Deep Ensemble     & $28.72_{\pm1.46}$ & $27.75_{\pm1.86}$ & $11.87_{\pm0.16}$ & $18.67_{\pm0.29}$ & $13.98_{\pm1.12}$ & $5.24_{\pm0.27}$ \\
& LA                & $11.41_{\pm0.17}$ & $30.54_{\pm0.70}$ & $45.85_{\pm2.08}$ & $10.80_{\pm0.38}$ & $35.65_{\pm1.14}$ & $18.22_{\pm0.41}$ \\
& BLoB              & $11.23_{\pm1.45}$ & $10.77_{\pm1.91}$ & $4.29_{\pm1.08}$  & $4.52_{\pm0.91}$  & $\second{3.82_{\pm0.96}}$  & $\second{1.46_{\pm0.36}}$ \\
& TFB               & $9.36_{\pm1.02}$ & $\second{7.37_{\pm0.21}}$ & $\best{3.03_{\pm0.43}}$ & $4.07_{\pm1.65}$ & $5.94_{\pm0.46}$ & $5.37_{\pm0.44}$ \\
& C-LoRA            & $\second{7.86_{\pm3.99}}$  & $8.83_{\pm1.20}$  & $4.27_{\pm1.24}$  & $\second{3.71_{\pm1.30}}$  & $4.00_{\pm0.84}$  & $1.62_{\pm0.44}$ \\
\cmidrule(lr){2-8}
\rowcolor{gray!15}
& DALorRA           & $\best{7.84_{\pm0.92}}$ & $\best{7.04_{\pm0.66}}$ & $\second{3.31_{\pm0.58}}$ & $\best{2.91_{\pm0.41}}$ & $\best{3.72_{\pm0.52}}$ & $\best{1.14_{\pm0.29}}$ \\
\midrule

\multirow{8}{*}{NLL (\downarrowcenter)} 
& MAP               & $2.86_{\pm0.23}$ & $3.07_{\pm0.09}$ & $1.13_{\pm0.10}$ & $1.26_{\pm0.12}$ & $1.04_{\pm0.02}$ & $0.34_{\pm0.00}$ \\
& MCD               & $2.50_{\pm0.12}$ & $2.81_{\pm0.25}$ & $1.13_{\pm0.04}$ & $1.16_{\pm0.03}$ & $1.01_{\pm0.07}$ & $\second{0.32_{\pm0.00}}$ \\
& Deep Ensemble     & $2.44_{\pm0.23}$ & $2.20_{\pm0.03}$ & $0.91_{\pm0.05}$ & $1.04_{\pm0.09}$ & $0.87_{\pm0.03}$ & $\second{0.32_{\pm0.00}}$ \\
& LA                & $\best{0.62_{\pm0.00}}$ & $1.17_{\pm0.01}$ & $0.97_{\pm0.05}$ & $0.56_{\pm0.00}$ & $0.98_{\pm0.01}$ & $0.45_{\pm0.00}$ \\
& BLoB              & $0.66_{\pm0.01}$ & $0.88_{\pm0.03}$ & $\second{0.44_{\pm0.00}}$ & $\second{0.54_{\pm0.00}}$ & $0.51_{\pm0.01}$ & $\best{0.31_{\pm0.01}}$ \\
& TFB               & $\second{0.62_{\pm0.03}}$ & $\second{0.86_{\pm0.01}}$ & $\best{0.42_{\pm0.03}}$ & $0.56_{\pm0.03}$ & $\second{0.50_{\pm0.01}}$ & $0.34_{\pm0.00}$ \\
& C-LoRA            & $0.63_{\pm0.02}$ & $0.88_{\pm0.00}$ & $0.48_{\pm0.02}$ & $0.57_{\pm0.03}$ & $0.59_{\pm0.05}$ & $0.35_{\pm0.02}$ \\
\cmidrule(lr){2-8}
\rowcolor{gray!15}
& DALorRA           & $0.62_{\pm0.02}$ & $\best{0.83_{\pm0.02}}$ & $0.45_{\pm0.01}$ & $\best{0.54_{\pm0.01}}$ & $\best{0.49_{\pm0.02}}$ & $\second{0.32_{\pm0.00}}$ \\
\bottomrule
\end{tabular}
}
\end{table*}

\subsection{Results on Combined Datasets}
\label{app:combined_datasets}
\begin{figure*}[!th]
    \centering
    \includegraphics[width=0.95\textwidth]{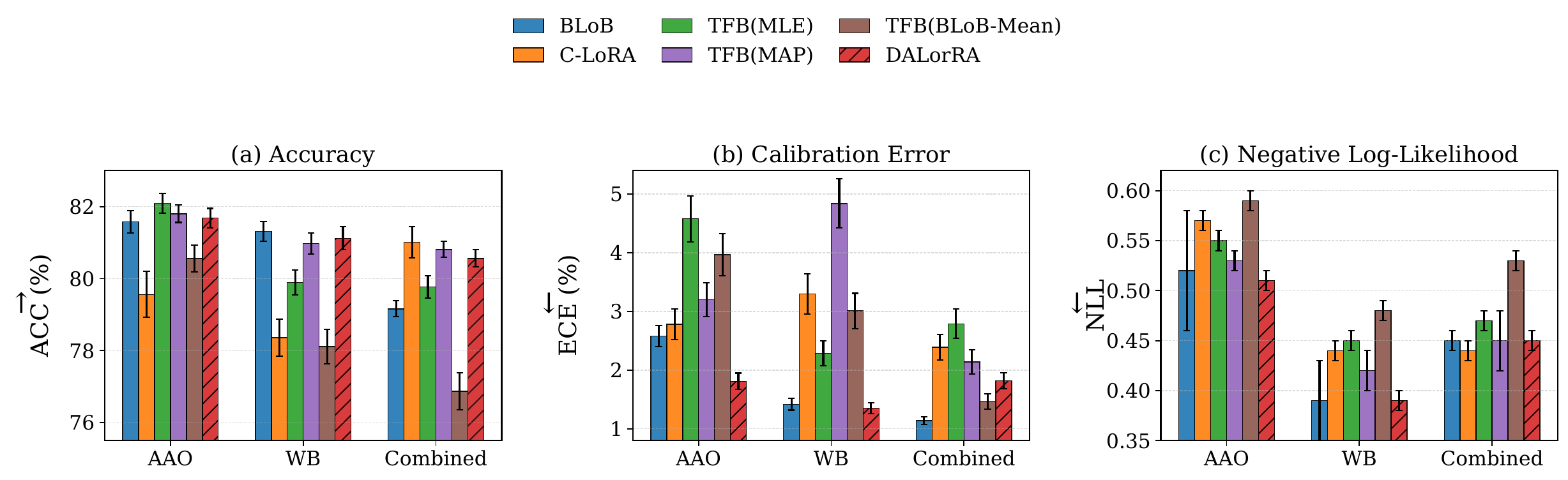}
    \caption{
    Performance comparison on combined datasets. We use $M = 5$ samples during inference in all sampling-based methods, including prior {BLoB}~\citep{wang2024blob}, {TFB}~\citep{shi2026training}, and {C-LoRA}~\citep{rahmati2026c}.
    AAO combines ARC-E, ARC-C, and OBQA; WB combines WG-M and BoolQ; and Combined merges all six benchmarks. 
    DALorRA achieves strong calibration on AAO and WB while maintaining competitive accuracy across merged settings.
    }
    \label{fig:combined_dataset_results}
\vspace{0.5cm}
\end{figure*}
Figure~\ref{fig:combined_dataset_results} reports the results on combined datasets, where related benchmarks are merged according to their answer-space compatibility. 
Compared with the single-dataset results in Table~\ref{tab:main_results_llama2_7b}, calibration metrics are generally improved on the merged datasets. 
In particular, ECE and NLL tend to decrease after merging related tasks, suggesting that increasing the effective training size may help alleviate overconfidence during LoRA fine-tuning.

Among the compared methods, our variants maintain competitive accuracy while achieving strong calibration performance. 
On the AAO and WB splits, our method achieves the lowest ECE, indicating that rank-level Bayesianization remains effective when related tasks are jointly trained. 
On the fully combined dataset, our DALorRA variant also obtains competitive ACC and NLL compared with BLoB, C-LoRA, and TFB variants.

These observations should not be interpreted as a strict causal conclusion, 
but they are consistent with prior findings that calibration can be affected by the effective training data scale and the amount of data available relative to model capacity~\citep{wang2024blob,cheon2026brain}.
Overall, the combined-dataset results suggest that DALorRA provides a favorable trade-off between predictive performance and calibration, while also indicating that data scale is an important factor in uncertainty estimation for LoRA-adapted LLMs.

\subsection{Effect of the  Prior Bernoulli Probability}
\label{app:prior_sensitivity}
\begin{figure*}[!th]
    \centering
    \includegraphics[width=0.95\textwidth]{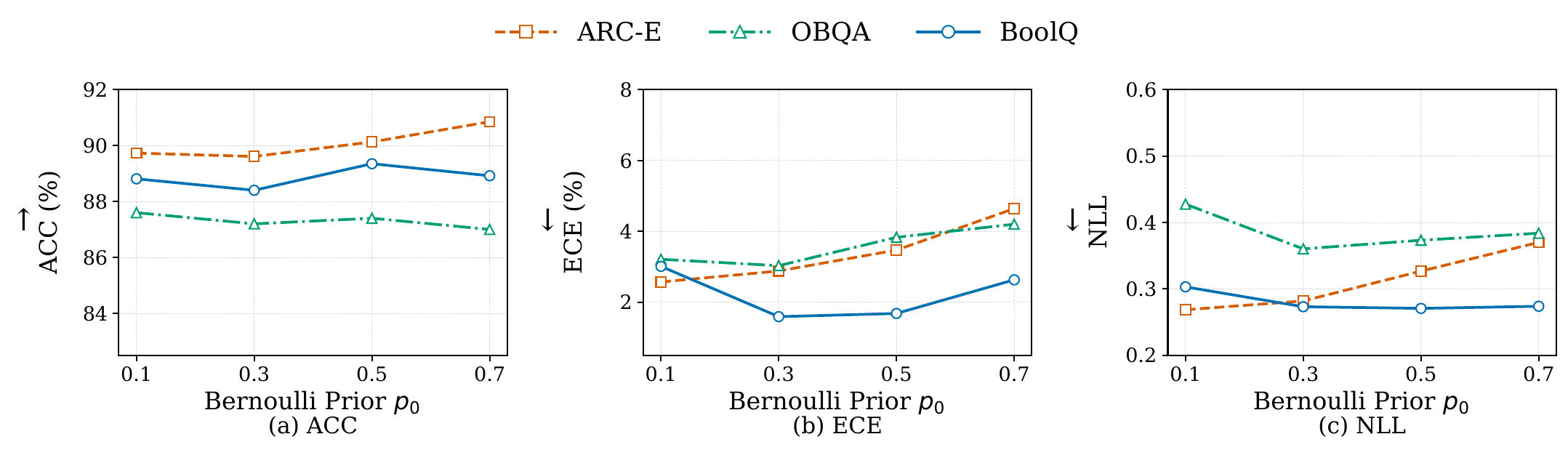}
    \caption{
        Sensitivity analysis of the prior Bernoulli  probability $p_0$
    }
    \label{fig:ablation_prior}
\end{figure*}
\begin{table*}[!th]
\centering
\caption{
{Efficiency Comparison among LoRA, BLoB, C-LoRA, and DALorRA.} The experiments are conducted on a single NVIDIA A40 GPU and based on Llama-3.1-8B with rank $r=8$, batch size 4, and 2,000 training iterations. The subscripts indicate the relative cost compared to LoRA,with \textcolor{red}{red} or \textcolor{green!50!black}{green} denoting worse or better efficiency.
}
\label{tab:efficiency_comparison}
\resizebox{\textwidth}{!}{
\begin{tabular}{lcccccccc}
\toprule
\multirow{2}{*}{Method} 
& \multirow{2}{*}{Trainable/Extra Params}
& \multicolumn{2}{c}{ARC-C}
& \multicolumn{2}{c}{WG-M}
& \multicolumn{2}{c}{BoolQ} \\
\cmidrule(lr){3-4} \cmidrule(lr){5-6} \cmidrule(lr){7-8}
& 
& Train/Eval (s) & Mem. (MB)
& Train/Eval (s) & Mem. (MB)
& Train/Eval (s) & Mem. (MB) \\
\midrule

LoRA 
& $4{,}466{,}688/0$
& $1{,}977.63_{\textcolor{black}{1.00\times}}/38.28_{\textcolor{black}{1.00\times}}$
& $11{,}462.94_{\textcolor{black}{1.00\times}}$
& $1{,}266.59_{\textcolor{black}{1.00\times}}/105.79_{\textcolor{black}{1.00\times}}$
& $7{,}755.07_{\textcolor{black}{1.00\times}}$
& $3{,}530.78_{\textcolor{black}{1.00\times}}/756.36_{\textcolor{black}{1.00\times}}$
& $14{,}279.04_{\textcolor{black}{1.00\times}}$ \\

BLoB
& $6{,}596{,}608_{\textcolor{red}{1.48\times}}/+2{,}129{,}920$
& $2{,}034.89_{\textcolor{red}{1.03\times}}/383.57_{\textcolor{red}{10.02\times}}$
& $13{,}092.71_{\textcolor{red}{1.14\times}}$
& $1{,}316.69_{\textcolor{red}{1.04\times}}/1{,}046.98_{\textcolor{red}{9.90\times}}$
& $8{,}306.96_{\textcolor{red}{1.07\times}}$
& $3{,}600.76_{\textcolor{red}{1.02\times}}/7{,}581.51_{\textcolor{red}{10.02\times}}$
& $16{,}708.24_{\textcolor{red}{1.17\times}}$ \\

C-LoRA
& $5{,}044{,}928_{\textcolor{red}{1.13\times}}/+578{,}240$
& $5{,}015.91_{\textcolor{red}{2.54\times}}/194.37_{\textcolor{red}{5.08\times}}$
& $16{,}283.28_{\textcolor{red}{1.42\times}}$
& $4{,}676.12_{\textcolor{red}{3.69\times}}/738.95_{\textcolor{red}{6.99\times}}$
& $11{,}338.00_{\textcolor{red}{1.46\times}}$
& $5{,}717.40_{\textcolor{red}{1.62\times}}/2{,}684.16_{\textcolor{red}{3.55\times}}$
& $20{,}601.88_{\textcolor{red}{1.44\times}}$ \\

DALorRA (Ours)
& $4{,}467{,}208_{\textcolor{red}{1.0001\times}}/+520$
& $1{,}125.13_{\textcolor{green!60!black}{0.57\times}}/223.63_{\textcolor{red}{5.84\times}}$
& $13{,}880.68_{\textcolor{red}{1.21\times}}$
& $895.07_{\textcolor{green!60!black}{0.71\times}}/810.41_{\textcolor{red}{7.66\times}}$
& $9{,}216.75_{\textcolor{red}{1.19\times}}$
& $1{,}640.12_{\textcolor{green!60!black}{0.46\times}}/3{,}321.28_{\textcolor{red}{4.39\times}}$
& $17{,}398.39_{\textcolor{red}{1.22\times}}$ \\

\bottomrule
\end{tabular}
}
\end{table*}
\begin{figure}[t]
    \centering
    \includegraphics[width=1\columnwidth]{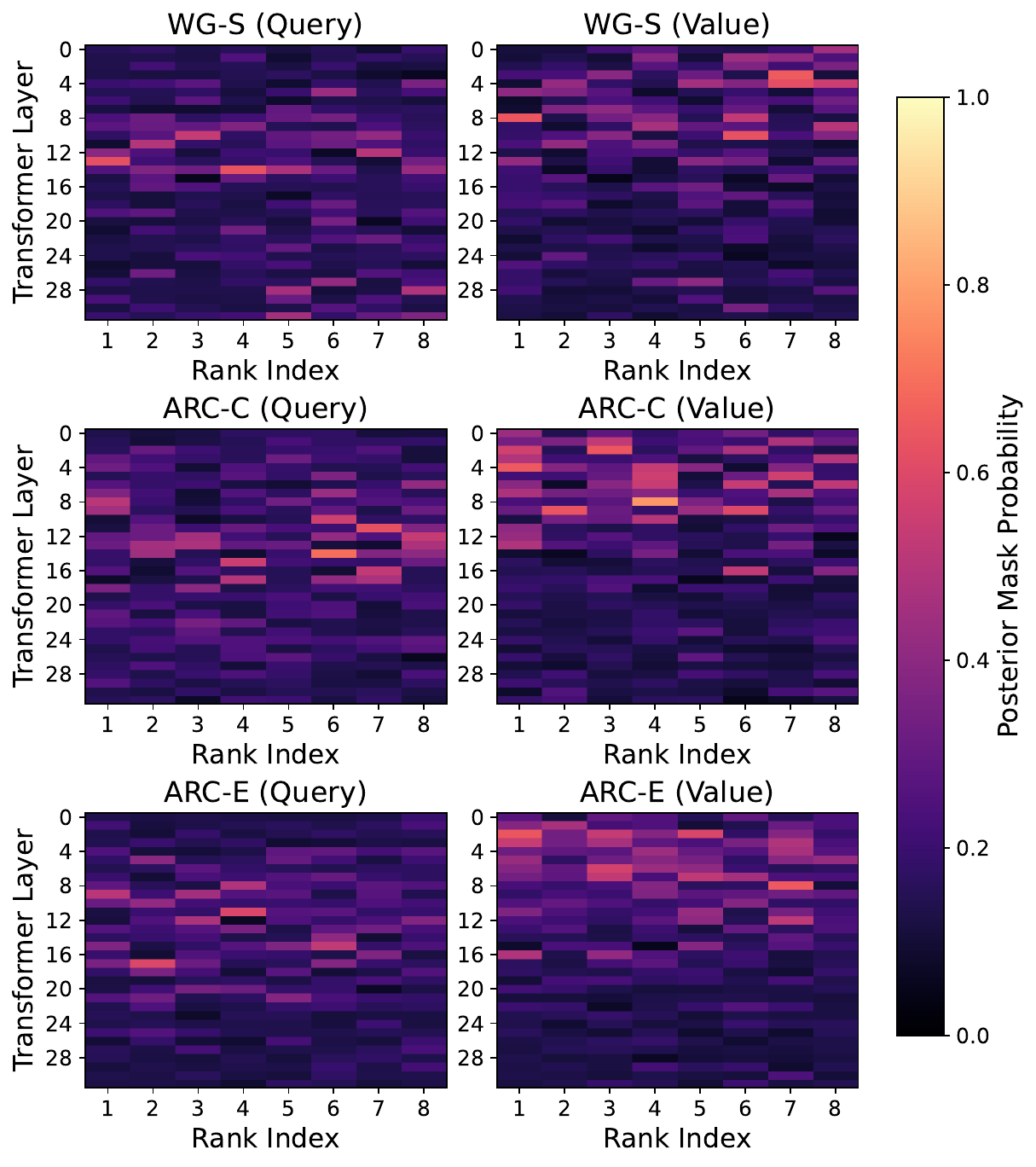}
    \caption{
    Posterior Bernoulli probabilities of DALorRA masks $\mathbf{D}$ on WG-S, ARC-C, and ARC-E.
    }
    \label{fig:rank_mask_heatmap_others}
\end{figure}

We study the impact of the prior Bernoulli probability, $p_0$, on the performance of DALorRA. As illustrated in Figure~\ref{fig:ablation_prior}, the performance is robust when $p_0$ varies within a reasonable range. The small fluctuation in performance with $p_0$ is theoretically expected, as the prior influences the regularization intensity of the Kullback-Leibler divergence. 

\subsection{Efficiency and resource consumption}
\label{app:resource consumption}

We compare the efficiency of the sampling-based uncertainty-aware fine-tuning methods (BLoB, C-LoRA, and DALorRA) with regular LoRA.
As shown in Table~\ref{tab:efficiency_comparison}, DALorRA adds only 520 extra trainable parameters, which is negligible compared with BLoB and C-LoRA. 
DALorRA takes the least training time, even faster than LoRA due to its sparse configuration; it uses only $0.57\times$, $0.71\times$, and $0.46\times$ the training time of LoRA on ARC-C, WG-M, and BoolQ, respectively.
DALorRA remains faster than BLoB and comparable to C-LoRA in evaluation (inference) time with moderate memory overhead. 
In addition, DALorRA's overall (training and evaluation) time is comparable to LoRA's on the three datasets.
This shows that DALorRA achieves great calibration with minimal computational cost and favorable overall efficiency.

\begin{table*}[!t]
\centering
\caption{Performance of Bayesian LoRA methods on Llama-3.1-8B under deterministic posterior-mean inference, without Monte Carlo sampling.}
\label{tab:mean_inference_id}
\footnotesize
\setlength{\tabcolsep}{3.5pt}
\renewcommand{\arraystretch}{0.95}
\resizebox{\textwidth}{!}{%
\begin{tabular}{@{}llcccccccccc@{}}
\toprule
& & \multicolumn{6}{c}{In-Distribution}
& \multicolumn{4}{c}{Out-of-Distribution (OBQA$\rightarrow$X)} \\
\cmidrule(lr){3-8}
\cmidrule(lr){9-12}

& & \multicolumn{6}{c}{}
& \multicolumn{2}{c}{\textit{Small Shift}}
& \multicolumn{2}{c}{\textit{Large Shift}} \\
\cmidrule(lr){9-10}
\cmidrule(lr){11-12}

Metric & Method
& WG-S & ARC-C & ARC-E & WG-M & OBQA & BoolQ
& ARC-C & ARC-E & Chem & Phy \\
\midrule

\multirow{4}{*}{ACC ($\uparrow$)}
& CLoRA
& 73.77$\pm$3.41
& 81.76$\pm$0.48
& \textbf{90.32$\pm$0.75}
& 82.96$\pm$1.01
& 86.10$\pm$1.56
& 88.86$\pm$0.48
& 78.56$\pm$0.33
& 86.20$\pm$1.24
& 43.56$\pm$0.13
& 44.23$\pm$0.22 \\

& BLoB
& \textbf{78.09$\pm$0.34}
& \textbf{83.11$\pm$2.39}
& 90.25$\pm$0.50
& 83.15$\pm$0.22
& 87.90$\pm$0.42
& \textbf{89.43$\pm$0.11}
& 80.21$\pm$0.00
& 86.61$\pm$0.54
& \textbf{45.31$\pm$1.56}
& 45.83$\pm$2.08 \\

& TFB
& 77.89$\pm$0.06
& 80.74$\pm$0.48
& 90.25$\pm$0.50
& 83.15$\pm$0.22
& \textbf{88.10$\pm$0.22}
& 89.12$\pm$0.15
& 79.86$\pm$0.35
& \textbf{86.96$\pm$0.36}
& 43.75$\pm$1.04
& \textbf{46.35$\pm$0.52}\\

& Ours
& 77.89$\pm$0.45
& 82.12$\pm$0.24
& 89.73$\pm$0.12
& \textbf{83.57$\pm$0.28}
& 87.80$\pm$0.18
& 89.24$\pm$0.12
& \textbf{80.24$\pm$0.22}
& 86.46$\pm$0.11
& 44.33$\pm$0.14
& 45.86$\pm$0.38\\
\midrule

\multirow{4}{*}{ECE ($\downarrow$)}
& CLoRA
& 21.89$\pm$2.26
& 13.29$\pm$0.16
& 6.60$\pm$1.62
& 11.98$\pm$1.24
& 7.00$\pm$1.14
& 5.02$\pm$1.09
& 9.22$\pm$1.26
& 6.12$\pm$0.34
& 19.34$\pm$0.44
& 20.64$\pm$0.22\\

& BLoB
& 14.83$\pm$0.53
& \textbf{10.48$\pm$1.19}
& 4.09$\pm$0.48
& 8.14$\pm$1.14
& 5.04$\pm$0.59
& 3.56$\pm$0.79
& 8.28$\pm$0.10
& \textbf{4.71$\pm$0.16}
& \textbf{17.98$\pm$1.32}
& 20.57$\pm$0.17 \\

& TFB
& 14.87$\pm$0.67
& 11.64$\pm$0.47
& \textbf{3.98$\pm$0.46}
& 8.14$\pm$1.13
& 5.06$\pm$0.19
& 4.30$\pm$0.06
& \textbf{8.00$\pm$0.29}
& 4.90$\pm$0.58
& 22.08$\pm$1.07
& \textbf{18.01$\pm$1.06} \\

& Ours
& \textbf{14.02$\pm$0.56}
& 11.06$\pm$0.47
& 4.62$\pm$0.23
& \textbf{7.69$\pm$0.29}
& \textbf{3.83$\pm$0.14}
& \textbf{3.25$\pm$0.16}
& \textbf{8.00$\pm$0.11}
& 5.12$\pm$0.36
& 19.66$\pm$1.21
& 18.24$\pm$0.34 \\
\midrule

\multirow{4}{*}{NLL ($\downarrow$)}
& CLoRA
& 1.21$\pm$0.02
& 0.79$\pm$0.03
& 0.42$\pm$0.15
& 0.61$\pm$0.09
& 0.42$\pm$0.05
& 0.31$\pm$0.01
& 0.69$\pm$0.02
& 0.45$\pm$0.03
& \textbf{1.21$\pm$0.04}
& 1.37$\pm$0.01 \\

& BLoB
& \textbf{0.70$\pm$0.04}
& \textbf{0.69$\pm$0.01}
& \textbf{0.27$\pm$0.02}
& 0.51$\pm$0.02
& 0.36$\pm$0.02
& 0.27$\pm$0.01
& 0.62$\pm$0.01
& \textbf{0.40$\pm$0.02}
& 1.29$\pm$0.01
& 1.36$\pm$0.00\\

& TFB
& 0.72$\pm$0.01
& 0.75$\pm$0.02
& 0.28$\pm$0.02
& \textbf{0.50$\pm$0.02}
& 0.36$\pm$0.02
& 0.28$\pm$0.00
& \textbf{0.61$\pm$0.01}
& 0.41$\pm$0.00
& 1.35$\pm$0.03
& \textbf{1.34$\pm$0.04} \\

& Ours
& 0.73$\pm$0.02
& \textbf{0.69$\pm$0.01}
& 0.30$\pm$0.02
& \textbf{0.50$\pm$0.01}
& \textbf{0.35$\pm$0.01}
& \textbf{0.26$\pm$0.02}
& 0.64$\pm$0.01
& \textbf{0.40$\pm$0.02}
& 1.22$\pm$0.03
& 1.38$\pm$0.01 \\

\bottomrule
\end{tabular}%
}
\end{table*}

\subsection{Deterministic Posterior-Mean Inference}
Monte Carlo (MC) averaging requires multiple forward passes at inference. We therefore evaluate a deterministic variant of DALorRA: each Bernoulli rank mask is replaced by its learned posterior mean at test time, following the test-time scaling idea in dropout \citep{srivastava2014dropout}. This removes MC sampling and uses a single forward pass. We apply the same posterior-mean inference to BLoB, C-LoRA, and TFB for a fair comparison.

Table~\ref{tab:mean_inference_id} shows that DALorRA remains competitive without MC sampling. On the six in-distribution datasets, it achieves the lowest ECE on four datasets and the lowest or tied-lowest NLL on four datasets, while maintaining comparable accuracy. MC averaging can still improve accuracy by combining multiple rank-mask configurations, whereas deterministic inference provides a lower-latency alternative. Uncertainty distillation is another way to reduce sampling cost, but requires training an additional student model \citep{vejendla2025efficient}.

\end{document}